# iNavFIter: Next-Generation Inertial Navigation Computation Based on Functional Iteration

Yuanxin Wu, *Senior Member, IEEE*

*Abstract*— **Inertial navigation computation is to acquire the attitude, velocity and position information of a moving body by integrating inertial measurements from gyroscopes and accelerometers. Over half a century has witnessed great efforts in coping with the motion non-commutativity errors to accurately compute the navigation information as far as possible, so as not to compromise the quality measurements of inertial sensors. Highly dynamic applications and the forthcoming cold-atom precision inertial navigation systems demand for even more accurate inertial navigation computation. The paper gives birth to an inertial navigation algorithm to fulfill that demand, named the iNavFIter, which is based on a brand-new framework of functional iterative integration and Chebyshev polynomials. Remarkably, the proposed iNavFIter reduces the non-commutativity errors to almost machine precision, namely, the coning/sculling/scrolling errors that have perplexed the navigation community for long. Numerical results are provided to demonstrate its accuracy superiority over the state-of-the-art inertial navigation algorithms at affordable computation cost.**

*Index Terms*—Inertial navigation, Navigation computation, Non-commutativity error, Picard integration, Chebyshev polynomial

## I. Introduction

Acquiring the attitude, velocity and position information is fundamental to any motion body manipulation. Inertial navigation is a dead-reckoning self-contained method to achieve this goal by integrating inertial measurements from two triads of gyroscopes and accelerometers. Over half a century has witnessed tremendous effects in fabricating inertial sensors with even further improved performance, as well as in designing advanced inertial navigation algorithm for strapdown systems so as not to compromise the quality of inertial sensors [1-4]. There is a popular belief that the modern inertial navigation algorithm has been good enough already for practical applications [7]. However, dynamic applications like military projectiles with high-speed complex rotation and the cold-atom interference gyroscopes of ultra-high precision (meters or less per hour in position error) on the horizon demand for even more accurate inertial navigation algorithm.

This work was supported in part by National Natural Science Foundation of China (61673263) and National Key R&D Program of China (2018YFB1305103). A short version was presented at 26th International Conference on Integrated Navigation Systems, Saint Petersburg, Russia, 2019; the 56th symposium on Inertial Sensors and Systems (ISS), Braunschweig, Germany, 2019.

Authors' address: Yuanxin Wu, Shanghai Key Laboratory of Navigation and Location-based Services, School of Electronic Information and Electrical Engineering, Shanghai Jiao Tong University, Shanghai, China, 200240, E-mail: (yuanx_wu@hotmail.com).

The computational cornerstone of modern-day strapdown inertial navigation lies in the coning/sculling/scrolling correction algorithms for integrating attitude/velocity/position differential equations [3, 9, 10]. For instance, the attitude algorithm structure, established in 1970s by Jordan [11] and Bortz [12], relies on the simplified rotation vector differential equation for incremental attitude update [9, 13-15]; the velocity algorithm takes the first-order approximation of the body attitude in the transformed specific force integration [6, 16]. The inertial navigation problem is represented in three dual quaternion differential quaternions (of the similar form with the attitude quaternion) and the coning correction-like screw algorithm is then developed to solve them [17]. Soon after in [18], the so-called velocity/position translation vectors, similar in spirit to the rotation vector, are devised and meant to be a design base for potential strapdown navigation algorithm development and assessment. The connection between the screw vectors in [17] and the velocity/position translation vectors is explored in [19-21]. A recent work [8] remedies the issue of reference frame rotation less-seriously treated in the usual velocity/position algorithms [6, 9]. Other attempts to further improve the modern-day navigation algorithm include the frequency-domain approach [22], using multiple integrals of angular rate/specific force [23] and considering the gyroscope frequency response effect [24]. In short, the state-of-the-art numerical navigation algorithm possesses two essential shortcomings. One is the theoretical simplification and approximation as stated above in handling the noncommutativity terms. The other shortcoming is that the state-of-the-art algorithm is optimized under specialized motion forms, e.g., the coning/sculling motions [6, 10, 14-16, 25, 26], and thus might likely be suboptimal under practical motions.

Quite recently, a significant advance has been made in precision attitude computation by independent groups [27-31]. They share the same spirit of trying to accurately solve the attitude kinematic equation based on the fitted angular velocity polynomial function. The main difference among these works is the chosen attitude parameterization. The quaternion is employed in [29], which first demonstrated in the public literature the practical potential of the Picard-type successive approximation method. The direction cosine matrix (DCM) is used instead by the Taylor series expansion [32]. These methods could be traced back to the Russian seminal work in 1990s [33], where they were respectively classified as Type I and Type II methods. In specific, the Type II method used the rotation vector instead of DCM. The three-component attitude parameterization is minimal and does not need to satisfy the inherent constraints of the redundant-component parameterizations, such as the quaternion with the unit-norm constraint and the DCM with the orthogonal and +1-determinant constraints. In view of the finite-polynomial-like differential equation of the three-component Rodrigues vector, Wu [27] proposes the RodFIter method to reconstruct the attitude, which makes a natural use of the functional iteration integration of the Rodrigues vector's kinematic equation. It highlights the capability of analytical attitude reconstruction over the whole update interval and provably converges to the true attitude if only the angular velocity is exact. Unfortunately, all high-accurate attitude algorithms face the problem of high computational burden, especially for real-time applications. Hence, to improve the computational efficiency, Wu [28] further brings forth a substantially fast version of RodFIter at little expense of accuracy by exploiting the excellent property of the Chebyshev polynomial. It reformulates the original RodFIter in terms of the iterative computation of the Rodrigues vector's Chebyshev polynomial coefficients and exerts Chebyshev polynomial truncation. In principle, the idea of RodFIter could be extended to various attitude parameters including the quaternion and the rotation vector (see the RotFIter addressed in [27] as well), but a question worthy of being investigated is whether the unit-norm constraint of quaternion affects the accuracy to be achieved [34-39]. It is right this concern that has prevented one from using quaternion for attitude computation before Yan's work [29]. The work [31] endeavoring to more accurately compute the rotation vector is in essence an example of RotFIter [27] with explicit but tedious iteration steps. Wu and Yan [30] raise the QuatFIter for attitude reconstruction using the quaternion in view of its linear kinematic equation. It is shown that the QuatFIter is equivalent to the previous Picard-type successive approximate quaternion method [33] and has about two times better computational efficiency at comparable accuracy to the RodFIter, although the latter has relatively uniform and faster error reduction with respect to the number of iterations. It is suspected therein that the unity-norm constraint of the quaternion contributes to QuatFIter's non-uniform error reduction, but no apparent relationship between the attitude error and the quaternion norm error has been identified so far.

The functional iterative integration combined with Chebyshev polynomial approximation was developed independently in the navigation community, but lately found to closely resemble the so-called Picard-Chebyshev method that was dated back to as early as 1960s [40]. In 1980s-1990s and even quite recently, it was employed and advanced by researchers in the field of astrodynamics for orbital determination [41-44].

The contribution of the paper is multiple-fold. The technique of Functional Iterative integration with Chebyshev polynomial approximation is successfully extended to the whole process of inertial Navigation computation (named the iNavFIter hereafter). Astonishingly, the non-commutativity (coning/sculling/scrolling) errors, which has long perplexed the navigation community for over half a century, is utterly eliminated to almost the machine precision at affordable computation burden. It is believed that this work has set a solid algorithmic foundation for both highly-dynamic applications and the forthcoming next-generation ultra-precision inertial navigation systems. The remaining of the paper is organized as follows. Section II gives an overview of the basic navigation kinematics and discusses the selection of the computation reference frame. Section III presents implementation details of the iNavFIter, including the angular velocity/specific force fitting by Chebyshev polynomials and the attitude, velocity and position computation using the functional iterative integration with appropriately truncated Chebyshev polynomials. Section IV discusses the iNavFIter's relation to the typical modern-day inertial navigation algorithms. Section V is devoted to the convergence property and error characteristics. Section VI assesses the iNavFIter with simulation data and demonstrates its remarkable accuracy superiority over the state-of-the-art navigation algorithms. The conclusion is finally drawn in Section VII.

## II. Inertial Navigation and Choice of Computation Reference Frame

Denote by $N$ the navigation frame, by $B$ the inertial measurement unit's body frame, by $I$ the inertially non-rotating frame and by *E* the Earth frame. Any frame could be chosen as the navigation frame, but the commonly used reference frame of inertial navigation computation is the Earth frame and the local-level frame. Without the loss of generality, the local-level navigation frame in this paper takes the definition of North-Up-East [8, 45]. This paper uses the conventional symbol denotation in the navigation community. For an attitude matrix or quaternion describing the attitude of a target frame relative to a source frame (alternatively the rotation from a source frame to a target frame), the source frame appears as the subscript and the target frame appears as the superscript. A source frame and a target frame usually appear as a pair in the subscript of a vector, while the superscript of a vector denotes the frame the vector is expressed.

In general, the navigation (attitude, velocity and position) rate equations in a general navigation frame are well known as [4, 5, 9]

$$\dot{\mathbf{q}}_n^b = \mathbf{q}_n^b \circ \boldsymbol{\omega}_{nb}^b / 2 = \left( \mathbf{q}_n^b \circ \boldsymbol{\omega}_{ib}^b - \boldsymbol{\omega}_{in}^n \circ \mathbf{q}_n^b \right) / 2 \tag{1}$$

$$\dot{\mathbf{v}}^n = \mathbf{C}_b^n \mathbf{f}^b - \left( 2\boldsymbol{\omega}_{ie}^n + \boldsymbol{\omega}_{en}^n \right) \times \mathbf{v}^n + \mathbf{g}^n \tag{2}$$

$$\dot{\mathbf{p}}^n = \mathbf{R}_c \mathbf{v}^n \tag{3}$$

where $\mathbf{q}_n^b$ or $\mathbf{C}_n^b$ denotes the attitude quaternion or the attitude matrix of the body frame relative to the navigation frame, $\boldsymbol{\omega}_{nb}^b = \boldsymbol{\omega}_{ib}^b - \mathbf{C}_n^b \boldsymbol{\omega}_{in}^n$ the body angular rate with respect to the navigation frame that is expressed in the body frame, $\boldsymbol{\omega}_{ib}^b$ the body angular rate measured by gyroscopes in the body frame, $\mathbf{v}^n = \begin{bmatrix} v_N & v_U & v_E \end{bmatrix}^T$ the velocity relative to the Earth (known as the ground velocity) that is expressed in the navigation frame, $\mathbf{f}^b$ the specific force measured by accelerometers in the body frame, $\boldsymbol{\omega}_{ie}^n = \begin{bmatrix} \Omega \cos L & \Omega \sin L & 0 \end{bmatrix}^T$ the Earth rotation rate with respect to the inertial frame that is expressed in the navigation frame, $\boldsymbol{\omega}_{en}^n = \begin{bmatrix} v_E / (R_E + h) & v_E \tan L / (R_E + h) & -v_N / (R_N + h) \end{bmatrix}^T$ the angular rate of the navigation frame with respect to the Earth frame that is expressed in the navigation frame, and $\mathbf{g}^n$ the gravity vector in the navigation frame. The $3 \times 3$ skew symmetric matrix $(\cdot \times)$ is defined so that the cross product satisfies $\mathbf{a} \times \mathbf{b} = (\mathbf{a} \times) \mathbf{b}$ for arbitrary two vectors. The position $\mathbf{p}^n = \begin{bmatrix} \lambda & L & h \end{bmatrix}^T$ is described by the angular orientation of the local-level frame relative to the Earth frame, commonly expressed as longitude $\lambda$, latitude *L* and the height above the Earth surface *h*. The local curvature matrix $\mathbf{R}_c$ is a function of the current position as

$$\mathbf{R}_c = \begin{bmatrix} 0 & 0 & \dfrac{1}{(R_E + h)\cos L} \\ \dfrac{1}{R_N + h} & 0 & 0 \\ 0 & 1 & 0 \end{bmatrix} \tag{4}$$

where $R_E$ and $R_N$ are respectively the transverse radius of curvature and the meridian radius of curvature of the reference ellipsoid, which depends on the current position as well. All the quantities above are functions of time and, if not explicitly stated, their time dependences are omitted for the sake of symbolic brevity.

Note that the attitude quaternion $\mathbf{q}$ can be represented as a four-dimensional column vector of unit magnitude, i.e., $\mathbf{q} = \begin{bmatrix} s & \mathbf{\eta}^T \end{bmatrix}^T$, where $s$ is the scalar part and $\mathbf{\eta}$ is the vector part. If these two parts are regarded as a scalar quaternion and a vector quaternion, respectively, then quaternion can be alternatively written as $\mathbf{q} = s + \mathbf{\eta}$. With some abuse of symbols, a vector quaternion is taken equally as a three-dimensional column vector throughout the paper. The operator $\circ$ in (1) means the multiplication of quaternions that is defined as

$$\mathbf{q}_1 \circ \mathbf{q}_2 = \begin{bmatrix} \mathbf{q}_1 \end{bmatrix}^+ \begin{bmatrix} s_2 \\ \mathbf{\eta}_2 \end{bmatrix} = \begin{bmatrix} \mathbf{q}_2 \end{bmatrix}^- \begin{bmatrix} s_1 \\ \mathbf{\eta}_1 \end{bmatrix} \tag{5}$$

The two quaternion multiplication matrices, $\begin{bmatrix} \mathbf{q} \end{bmatrix}^+$ and $\begin{bmatrix} \mathbf{q} \end{bmatrix}^-$, are respectively defined by

$$\begin{bmatrix} \mathbf{q} \end{bmatrix}^+ = \begin{bmatrix} s & -\mathbf{\eta}^T \\ \mathbf{\eta} & s\,\mathbf{I}_3 + \mathbf{\eta} \times \end{bmatrix}, \quad \begin{bmatrix} \mathbf{q} \end{bmatrix}^- = \begin{bmatrix} s & -\mathbf{\eta}^T \\ \mathbf{\eta} & s\,\mathbf{I}_3 - \mathbf{\eta} \times \end{bmatrix} \tag{6}$$

The attitude matrix $\mathbf{C}_b^n$ is related to the attitude quaternion $\mathbf{q}_n^b = \begin{bmatrix} s & \mathbf{\eta}^T \end{bmatrix}^T$ by [4]

$$\mathbf{C}_b^n = \left( s^2 - \mathbf{\eta}^T \mathbf{\eta} \right) \mathbf{I}_3 + 2\mathbf{\eta}\mathbf{\eta}^T + 2s\mathbf{\eta} \times \tag{7}$$

When we let the navigation frame coincide with the Earth frame, the Earth frame navigation rate equations can be readily obtained from (1)-(3) as

$$\dot{\mathbf{q}}_e^b = \mathbf{q}_e^b \circ \mathbf{\omega}_{eb}^b / 2 = \left( \mathbf{q}_e^b \circ \mathbf{\omega}_{ib}^b - \mathbf{\omega}_{ie}^e \circ \mathbf{q}_e^b \right) / 2 \tag{8}$$

$$\dot{\mathbf{v}}^e = \mathbf{C}_b^e \mathbf{f}^b - 2\mathbf{\omega}_{ie}^e \times \mathbf{v}^e + \mathbf{g}^e \tag{9}$$

$$\dot{\mathbf{p}}^e = \mathbf{v}^e \tag{10}$$

where $\mathbf{p}^e = \begin{bmatrix} x & y & z \end{bmatrix}^T$ denotes the Earth-centered Earth-fixed (ECEF) coordinate, $\mathbf{v}^e = \begin{bmatrix} v_x & v_y & v_z \end{bmatrix}^T$ is the ground velocity expressed in the Earth frame and $\mathbf{\omega}_{ie}^e = \begin{bmatrix} 0 & 0 & \Omega \end{bmatrix}^T$ is the Earth rotation rate expressed in the Earth frame.

Figure 1 plots the information flow chart for both the local-level frame navigation equation and the Earth frame navigation equation. The attitude, velocity and position are tightly coupled in the local-level frame navigation equation (1)-(3). For instance, the attitude rate equation (1) needs to know the velocity and position information for computing $\mathbf{\omega}_{in}^n$. In contrast, the attitude, velocity and position are loosely-coupled in the Earth frame navigation equation in that the information flows unidirectionally from attitude (8), passing through velocity (9), to position (10). Note that there is a feedback from position to velocity, because the gravity vector is a function of position. In view of the functional iteration process below, the loosely-coupling effect is beneficial to reduce the computational cost, so we use the Earth frame as the computation reference frame in the sequel. The navigation information with respect to other frames, such as the local-level frame, could be readily obtained by appropriate transformation out of the dead-reckoning computation loop [4]. An additional benefit is that the Earth frame is singularity-free, in contrast to the local-level frame that would encounter a serious singular problem at polar areas [4, 5, 9, 46]. Specifically, the local curvature matrix $\mathbf{R}_c$ and the navigation frame's angular rate $\boldsymbol{\omega}_{en}^n$ will be subject to numerical problems while the latitude $L$ approaches $\pi/2$.

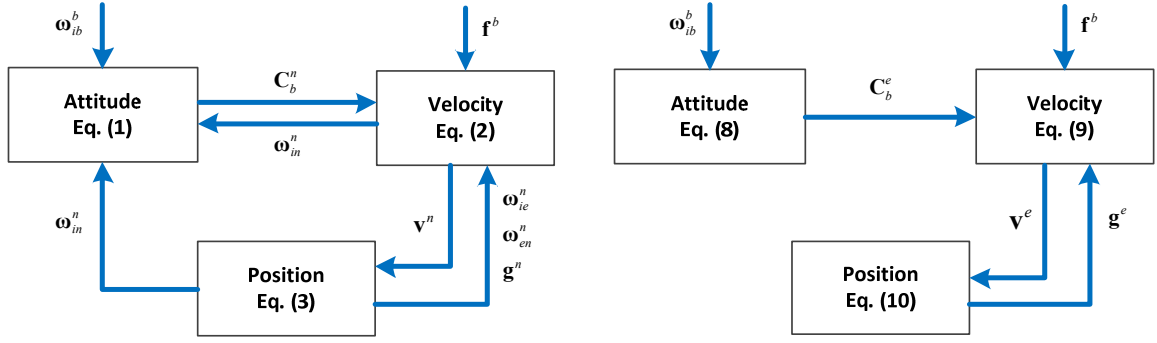

Figure 1. Information flow in local-level frame mechanization (left) and Earth-frame mechanization (right). Arrowed lines indicate information flow directions and the associated symbols mean that their computation needs to feed on the source information.

### III. Precision Navigation Computation Based on Functional Iterative Integration

Without the loss of generality, we will consider the navigation updates over the time interval $[0 \quad t]$, in which $N$ samples of triads of gyroscopes and accelerometers are available. For the sake of symbolic brevity as well, the E-frame navigation equations in (8)-(10) are simplified by omitting the redundant superscripts/subscripts as follows:

$$\dot{\mathbf{q}} = \left(\mathbf{q} \circ \boldsymbol{\omega}^b - \boldsymbol{\omega}_e \circ \mathbf{q}\right)\big/2 \tag{11}$$

$$\dot{\mathbf{v}}^e = \mathbf{C}_b^e \mathbf{f}^b - 2\boldsymbol{\omega}_e \times \mathbf{v}^e + \mathbf{g}^e \tag{12}$$

$$\dot{\mathbf{p}}^e = \mathbf{v}^e \tag{13}$$

Their integrals over the time interval of interest yield, respectively,

$$\mathbf{q} = \mathbf{q}(0) + \int_0^t \left(\mathbf{q} \circ \boldsymbol{\omega}^b - \boldsymbol{\omega}_e \circ \mathbf{q}\right)\big/2 \, dt \tag{14}$$

$$\mathbf{v}^e = \mathbf{v}^e(0) + \int_0^t \left(\mathbf{C}_b^e \mathbf{f}^b - 2\boldsymbol{\omega}_e \times \mathbf{v}^e + \mathbf{g}^e\right) dt \tag{15}$$

$$\mathbf{p}^e = \mathbf{p}^e(0) + \int_0^t \mathbf{v}^e \, dt \tag{16}$$

where $\mathbf{q}(0)$, $\mathbf{v}^e(0)$ and $\mathbf{p}^e(0)$ are the initial attitude, the initial velocity and the initial position at time zero, respectively. Next we will try to fit the gyroscope/accelerometer measurements using the Chebyshev polynomials and then solve the above integrations by functional iteration.

*A. Angular Velocity and Specific Force Fitted by Chebyshev Polynomial*

The Chebyshev polynomial of the first kind is defined over the interval $\begin{bmatrix} -1 & 1 \end{bmatrix}$ by the recurrence relation as

$$F_0(x) = 1, F_1(x) = x, F_{i+1}(x) = 2xF_i(x) - F_{i-1}(x) \quad for \ i \geq 1 \tag{17}$$

where $F_i(x)$ is the $i$th-degree Chebyshev polynomial of the first kind. For any $j, k \geq 0$, it satisfies the equality [47]

$$F_j(\tau)F_k(\tau) = \frac{1}{2}\left(F_{j+k}(\tau) + F_{|j-k|}(\tau)\right) \tag{18}$$

According to the integral property of the Chebyshev polynomial [47], we have

$$G_{i,[\tau_{k-1} \ \tau_k]} = \int_{\tau_{k-1}}^{\tau_k} F_i(\tau)d\tau = \begin{cases} \left(\dfrac{iF_{i+1}(\tau_k)}{i^2-1} - \dfrac{\tau_k F_i(\tau_k)}{i-1}\right) - \left(\dfrac{iF_{i+1}(\tau_{k-1})}{i^2-1} - \dfrac{\tau_{k-1}F_i(\tau_{k-1})}{i-1}\right), & i \neq 1 \\[2mm] \dfrac{\tau_k^2 - \tau_{k-1}^2}{2}, & i = 1 \end{cases} \tag{19}$$

In specific, the integrated $i$th-degree Chebyshev polynomial can be expressed as a linear combination of $(i+1)$th-degree Chebyshev polynomials, given by

$$G_{i,[-1 \ \tau]} = \int_{-1}^{\tau} F_i(\tau)d\tau = \begin{cases} \left(\dfrac{iF_{i+1}(\tau)}{i^2-1} - \dfrac{\tau F_i(\tau)}{i-1}\right) - \left(\dfrac{iF_{i+1}(-1)}{i^2-1} + \dfrac{F_i(-1)}{i-1}\right)F_0(\tau) \\[2mm] = \left(\dfrac{iF_{i+1}(\tau)}{i^2-1} - \dfrac{F_{i+1}(\tau) + F_{|i-1|}(\tau)}{2(i-1)}\right) - \left(\dfrac{iF_{i+1}(-1)}{i^2-1} + \dfrac{F_i(-1)}{i-1}\right)F_0(\tau) \\[2mm] = \left(\dfrac{F_{i+1}(\tau)}{2(i+1)} - \dfrac{F_{|i-1|}(\tau)}{2(i-1)}\right) - \dfrac{(-1)^i}{i^2-1}F_0(\tau) \quad \text{for } i \neq 1, \\[2mm] \dfrac{\tau^2-1}{2} = \dfrac{F_{i+1}(\tau)}{4} - \dfrac{F_0(\tau)}{4} \quad \text{for } i = 1, \end{cases} \tag{20}$$

At time instants $t_k$ $(k = 1, 2, \ldots N)$, assume discrete angular velocity $\tilde{\boldsymbol{\omega}}_{t_k}$ or angular increment (integrated angular velocity) $\Delta\tilde{\boldsymbol{\theta}}_{t_k}$ measurements by a triad of gyroscopes, and discrete specific force $\tilde{\mathbf{f}}_{t_k}$ or velocity increments (integrated specific force) $\Delta\tilde{\mathbf{v}}_{t_k}$ measurements by a triad of accelerometers. In order to apply the Chebyshev polynomials, the actual time interval is mapped onto $\begin{bmatrix} -1 & 1 \end{bmatrix}$ by letting $t = t_N(1+\tau)/2$. The fitted angular velocity and specific force using the Chebyshev polynomials can be respectively written as [27, 28]

$$\hat{\boldsymbol{\omega}}^b = \sum_{i=0}^{n_\omega} \mathbf{c}_i F_i(\tau), \quad n_\omega \leq N-1 \tag{21}$$

and

$$\hat{\mathbf{f}}^b = \sum_{i=0}^{n_f} \mathbf{d}_i F_i(\tau), \quad n_f \leq N-1 \tag{22}$$

where $n_\omega$ and $n_f$ denote the maximum degrees of Chebyshev polynomials. The coefficients $\mathbf{c}_i$ and $\mathbf{d}_i$ are determined by solving the least-square equations of the discrete gyroscope/accelerometer measurements. In specific, for the case of discrete angular velocity/specific force measurements, the coefficients satisfy

$$\boldsymbol{\Gamma}(n_\omega) \begin{bmatrix} \mathbf{c}_0 & \mathbf{c}_1 & \dots & \mathbf{c}_{n_\omega} \end{bmatrix}^T = \begin{bmatrix} \tilde{\boldsymbol{\omega}}_{t_1} & \tilde{\boldsymbol{\omega}}_{t_2} & \dots & \tilde{\boldsymbol{\omega}}_{t_N} \end{bmatrix}^T$$
$$\boldsymbol{\Gamma}(n_f) \begin{bmatrix} \mathbf{d}_0 & \mathbf{d}_1 & \dots & \mathbf{d}_{n_f} \end{bmatrix}^T = \begin{bmatrix} \tilde{\mathbf{f}}_{t_1} & \tilde{\mathbf{f}}_{t_2} & \dots & \tilde{\mathbf{f}}_{t_N} \end{bmatrix}^T \tag{23}$$

while for the case of discrete angular increment/velocity increment measurements, the coefficients satisfy

$$\Theta(n_\omega) \begin{bmatrix} \mathbf{c}_0 & \mathbf{c}_1 & \dots & \mathbf{c}_{n_\omega} \end{bmatrix}^T = \begin{bmatrix} \Delta\tilde{\boldsymbol{\theta}}_{t_1} & \Delta\tilde{\boldsymbol{\theta}}_{t_2} & \dots & \Delta\tilde{\boldsymbol{\theta}}_{t_N} \end{bmatrix}^T$$
$$\Theta(n_f) \begin{bmatrix} \mathbf{d}_0 & \mathbf{d}_1 & \dots & \mathbf{d}_{n_f} \end{bmatrix}^T = \begin{bmatrix} \Delta\tilde{\mathbf{v}}_{t_1} & \Delta\tilde{\mathbf{v}}_{t_2} & \dots & \Delta\tilde{\mathbf{v}}_{t_N} \end{bmatrix}^T \tag{24}$$

where the matrices $\boldsymbol{\Gamma}(n) = \begin{bmatrix} 1 & F_1(\tau_1) & \dots & F_n(\tau_1) \\ 1 & F_1(\tau_2) & \dots & F_n(\tau_2) \\ \vdots & \vdots & \vdots & \vdots \\ 1 & F_1(\tau_N) & \dots & F_n(\tau_N) \end{bmatrix}$ and $\Theta(n) = \frac{t_N}{2} \begin{bmatrix} G_{0,[\tau_0 \tau_1]} & G_{1,[\tau_0 \tau_1]} & \dots & G_{n,[\tau_0 \tau_1]} \\ G_{0,[\tau_1 \tau_2]} & G_{1,[\tau_1 \tau_2]} & \dots & G_{n,[\tau_1 \tau_2]} \\ \vdots & \vdots & \vdots & \vdots \\ G_{0,[\tau_{N-1} \tau_N]} & G_{1,[\tau_{N-1} \tau_N]} & \dots & G_{n,[\tau_{N-1} \tau_N]} \end{bmatrix}$.

If the angular velocity and the specific force are smooth, the fitted Chebyshev coefficients $\mathbf{c}_i$ and $\mathbf{d}_i$ will decrease exponentially in magnitude due to the orthogonal property of the Chebyshev polynomial [47].

## B. Attitude Computation

Hereafter we use the QuatFIter [30] to deal with attitude computation for the sake of its computational efficiency, although other attitude parameters could be alternatively employed, such as the Rodrigues vector in [27, 28]. Additionally, the quaternion has a simpler transformation to the attitude matrix than the Rodrigues vector does, which is preferable for the subsequent iterative computation of velocity.

Applying the functional iteration integration technique [27], the attitude quaternion in (14) can be iteratively computed as

$$\mathbf{q}_{l+1} = \mathbf{q}(0) + \frac{1}{2} \int_0^t \left( \mathbf{q}_l \circ \boldsymbol{\omega}^b - \boldsymbol{\omega}_e \circ \mathbf{q}_l \right) dt, \quad l \geq 0 \tag{25}$$

with some chosen initial attitude quaternion function, say $\mathbf{q}_0(t) \equiv \mathbf{q}(0)$. Assume the quaternion estimate at the $l$-th iteration is expressed by a weighted sum of Chebyshev polynomials, say

$$\mathbf{q}_l = \sum_{i=0}^{m_q} \mathbf{b}_{l,i} F_i(\tau) \tag{26}$$

where $m_q$ denotes the maximum degree and $\mathbf{b}_{l,i}$ is the coefficient of $i^{\text{th}}$-degree Chebyshev polynomial at the $l$-th iteration. The integral in (14) is transformed to that over the mapped interval of Chebyshev polynomials, that is,

$$\mathbf{q}_{l+1} = \mathbf{q}(0) + \frac{t_N}{4} \int_{-1}^{\tau} \left( \mathbf{q}_l \circ \boldsymbol{\omega}^b - \boldsymbol{\omega}_e \circ \mathbf{q}_l \right) d\tau \tag{27}$$

Substituting (21) and (26) and noting that the Earth rotation rate expressed in the Earth frame $\boldsymbol{\omega}_e$ is constant,

$$\begin{aligned}
\mathbf{q}_{l+1} &= \mathbf{q}(0) + \frac{t_N}{4} \int_{-1}^{\tau} \left( \left[ \sum_{i=0}^{m_q} \mathbf{b}_{l,i} F_i(\tau) \right] \circ \sum_{j=0}^{n_\omega} \mathbf{c}_j F_j(\tau) - \sum_{i=0}^{m_q} \boldsymbol{\omega}_e \circ \mathbf{b}_{l,i} F_i(\tau) \right) d\tau \\
&= \mathbf{q}(0) + \frac{t_N}{4} \left( \sum_{i=0}^{m_q} \sum_{j=0}^{n_\omega} \mathbf{b}_{l,i} \circ \mathbf{c}_j \int_{-1}^{\tau} F_i(\tau) F_j(\tau) d\tau - \sum_{i=0}^{m_q} \boldsymbol{\omega}_e \circ \mathbf{b}_{l,i} \int_{-1}^{\tau} F_i(\tau) d\tau \right) \\
&= \mathbf{q}(0) + \frac{t_N}{8} \left( \sum_{i=0}^{m_q} \sum_{j=0}^{n_\omega} \mathbf{b}_{l,i} \circ \mathbf{c}_j \left( G_{i+j,[-1\ \tau]} + G_{|i-j|,[-1\ \tau]} \right) - 2 \sum_{i=0}^{m_q} \boldsymbol{\omega}_e \circ \mathbf{b}_{l,i} G_{i,[-1\ \tau]} \right) \\
&= \sum_{i=0}^{m_q+n_\omega+1} \mathbf{b}_{l+1,i} F_i(\tau) \quad \overset{\text{polynomial truncation}}{\approx} \quad \sum_{i=0}^{m_q} \mathbf{b}_{l+1,i} F_i(\tau)
\end{aligned} \tag{28}$$

The last approximation is due to the polynomial truncation with a prescribed maximum degree, say $m_q$ as in (26), for improving computational efficiency [30]. As a quaternion multiplication involves 16 scalar multiplications, the computation complexity involved is roughly proportional to $\mathrm{O}\left(16 m_q n_\omega + 16 m_q\right)$ at each iteration. The iterative process could be repeated until some prescribed maximum iteration number was reached or some convergence criterion was met. For instance, the convergence criterion could be set as the root mean square (RMS) of the discrepancy of the polynomial coefficients between successive iterations, namely, $\left( \sum_{i=0}^{m_q} \left| \mathbf{b}_{l+1,i} - \mathbf{b}_{l,i} \right|^2 \right)^{1/2}$, is less than some threshold. The resultant quaternion of the iterative attitude computation in (28) is denoted by $\mathbf{q} = \sum_{i=0}^{m_q} \mathbf{b}_i F_i(\tau)$ to be used subsequently as an input.

*C. Velocity and Position Computation*

The gravity models in the literature are described either in the Earth frame or the local-level frame, but they are not exactly consistent with each other. Therefore, in this paper a gravity model in the local-level frame is used but expressed in the Earth frame accordingly, that is to say,

$$\mathbf{g}^e = \mathbf{C}_n^e \mathbf{g}^n \tag{29}$$

where the attitude matrix between the Earth frame and the local-level frame and the gravity are both functions of the current curvilinear position. In specific,

$$\mathbf{C}_n^e = \begin{bmatrix} -\sin L \cos \lambda & \cos L \cos \lambda & -\sin \lambda \\ -\sin L \sin \lambda & \cos L \sin \lambda & \cos \lambda \\ \cos L & \sin L & 0 \end{bmatrix} \text{ and } \mathbf{g}^n = \begin{bmatrix} 0 & -g & 0 \end{bmatrix}^T \tag{30}$$

where $g$ denotes the normal gravity in the vertical direction. In principle any normal gravity model could be applied if the computational burden involved was acceptable.

Applying the functional iteration integration technique [27] to (15)-(16), the velocity and position can be iteratively computed as

$$\mathbf{v}_{l+1}^e = \mathbf{v}^e(0) + \int_0^t \left[ \mathbf{C}_b^e(\mathbf{q}) \mathbf{f}^b - 2\boldsymbol{\omega}_e \times \mathbf{v}_l^e + \mathbf{g}^e(\mathbf{p}_l^e) \right] dt = \mathbf{v}^e(0) + \frac{t_N}{2} \int_{-1}^{\tau} \left[ \mathbf{C}_b^e(\mathbf{q}) \mathbf{f}^b - 2\boldsymbol{\omega}_e \times \mathbf{v}_l^e + \mathbf{g}^e(\mathbf{p}_l^e) \right] d\tau \tag{31}$$

and

$$\mathbf{p}_{l+1}^e = \mathbf{p}^e(0) + \int_0^t \mathbf{v}_l^e \, dt = \mathbf{p}^e(0) + \frac{t_N}{2} \int_{-1}^{\tau} \mathbf{v}_l^e \, d\tau \tag{32}$$

with some chosen initial velocity/position functions, say $\mathbf{v}_0^e(t) \equiv \mathbf{v}^e(0)$ and $\mathbf{p}_0^e(t) \equiv \mathbf{p}^e(0)$. Note that the attitude matrix $\mathbf{C}_b^e$ is obtained from the resultant quaternion of the above iterative attitude computation and the gravity $\mathbf{g}^e$ is implicitly dependent on the current position. Using the Chebyshev polynomial's property (18)-(20) and the fitted specific force (22), we have

$$\begin{aligned} \mathbf{I}_f(\tau) &= \int_{-1}^{\tau} \mathbf{C}_b^e \mathbf{f}^b d\tau = \int_{-1}^{\tau} \mathbf{q} \circ \mathbf{f}^b \circ \mathbf{q}^* d\tau = \int_{-1}^{\tau} \left( \sum_{i=0}^{m_q} \mathbf{b}_i F_i(\tau) \right) \circ \left( \sum_{j=0}^{n_f} \mathbf{d}_j F_j(\tau) \right) \circ \left( \sum_{k=0}^{m_q} \mathbf{b}_k^* F_k(\tau) \right) d\tau \\ &= \sum_{i=0}^{m_q} \sum_{j=0}^{n_f} \sum_{k=0}^{m_q} \mathbf{b}_i \circ \mathbf{d}_j \circ \mathbf{b}_k^* \int_{-1}^{\tau} F_i(\tau) F_j(\tau) F_k(\tau) d\tau \\ &= \frac{1}{2} \sum_{i=0}^{m_q} \sum_{j=0}^{n_f} \sum_{k=0}^{m_q} \mathbf{b}_i \circ \mathbf{d}_j \circ \mathbf{b}_k^* \int_{-1}^{\tau} \left( F_{i+j}(\tau) + F_{|i-j|}(\tau) \right) F_k(\tau) d\tau \\ &= \frac{1}{4} \sum_{i=0}^{m_q} \sum_{j=0}^{n_f} \sum_{k=0}^{m_q} \mathbf{b}_i \circ \mathbf{d}_j \circ \mathbf{b}_k^* \int_{-1}^{\tau} \left( F_{i+j+k}(\tau) + F_{|i+j-k|}(\tau) + F_{|i-j|+k}(\tau) + F_{|i-j|-k|}(\tau) \right) d\tau \\ &= \frac{1}{4} \sum_{i=0}^{m_q} \sum_{j=0}^{n_f} \sum_{k=0}^{m_q} \mathbf{b}_i \circ \mathbf{d}_j \circ \mathbf{b}_k^* \left( G_{i+j+k,[-1\ \tau]} + G_{|i+j-k|,[-1\ \tau]} + G_{|i-j|+k,[-1\ \tau]} + G_{|i-j|-k|,[-1\ \tau]} \right) \end{aligned} \tag{33}$$

As evidenced in the QuatFIter [30], the norm of the quaternion estimate (28) approaches unity after sufficient iterations, so it is unnecessary to do the quaternion normalization in (33). This is a nice advantage of the QuatFIter. If other attitude parameters were used, e.g., the Rodrigues vector in the RodFIter [27, 28], the nonlinear transformation from these parameters to the attitude matrix would produce infinite polynomials, for which an additional polynomial approximation step has to be incorporated.

Assume the velocity/position estimates at the $l$-th iteration is given by weighted sums of Chebyshev polynomials, say

$$\mathbf{v}_l^e = \sum_{i=0}^{m_v} \mathbf{s}_{l,i} F_i(\tau), \quad \mathbf{p}_l^e = \sum_{i=0}^{m_p} \boldsymbol{\rho}_{l,i} F_i(\tau) \tag{34}$$

where $m_v$ and $m_p$ are the maximum degrees, and $\mathbf{s}_{l,i}$ and $\boldsymbol{\rho}_{l,i}$ are the coefficients of $i^{\text{th}}$-degree Chebyshev polynomials at the $l$-th iteration for velocity and position, respectively. The gravity in the Earth frame is a nonlinear function of the curvilinear position (see Appendix for an example of the WGS-84 gravity model) and yet can be approximated by a weighted sum of Chebyshev polynomials as such

$$\mathbf{g}^e\left(\mathbf{p}_l^e\right) = \mathbf{C}_n^e \mathbf{g}^n\left(\mathbf{p}_l^n\right) = \mathbf{C}_n^e \mathbf{g}^n\left(ecef2lla\left(\mathbf{p}_l^e\right)\right) = \mathbf{C}_n^e \mathbf{g}^n\left(ecef2lla\left(\sum_{i=0}^{m_p}\boldsymbol{\rho}_{l,i}F_i\left(\tau\right)\right)\right) \approx \sum_{i=0}^{m_g}\boldsymbol{\gamma}_{l,i}F_i\left(\tau\right) \tag{35}$$

where $m_g$ is the maximum degree. The function $ecef2lla(\cdot)$ means the coordinate transformation from the ECEF coordinate to the curvilinear coordinate, see e.g. [48-50]. The coefficient $\boldsymbol{\gamma}_{l,i}$ is approximately calculated by [47]

$$\begin{aligned}\boldsymbol{\gamma}_{l,i} &\approx \frac{2-\delta_{0i}}{P}\sum_{k=0}^{P-1}\cos\frac{i\left(k+1/2\right)\pi}{P}\mathbf{g}^e\left(\mathbf{p}_l^e\left(\cos\frac{\left(k+1/2\right)\pi}{P}\right)\right)\\ &= \frac{2-\delta_{0i}}{P}\sum_{k=0}^{P-1}\cos\frac{i\left(k+1/2\right)\pi}{P}\mathbf{C}_n^e\mathbf{g}^n\left(ecef2lla\left(\sum_{i=0}^{m_p}\boldsymbol{\rho}_{l,i}F_i\left(\cos\frac{\left(k+1/2\right)\pi}{P}\right)\right)\right)\end{aligned} \tag{36}$$

where $\delta_{0i}$ is the Kronecker delta function, yielding 1 for $i=1$ and zero otherwise. Exact coefficients could be obtained only if the number of summation terms $P$ approaches infinity.

Substitute (33)-(35) into (31)-(32), the iterative velocity/position computation become

$$\begin{aligned}\mathbf{v}_{l+1}^e &= \mathbf{v}^e\left(0\right) + \frac{t_N}{2}\left[\mathbf{I}_f - 2\int_{-1}^{\tau}\left(\boldsymbol{\omega}_e\times\sum_{i=0}^{m_v}\mathbf{s}_{l,i}F_i\left(\tau\right)\right)d\tau + \int_{-1}^{\tau}\sum_{i=0}^{m_g}\boldsymbol{\gamma}_{l,i}F_i\left(\tau\right)d\tau\right]\\ &= \mathbf{v}^e\left(0\right) + \frac{t_N}{2}\left(\mathbf{I}_f - 2\sum_{i=0}^{m_v}\boldsymbol{\omega}_e\times\mathbf{s}_{l,i}G_{i,[-1\ \tau]} + \sum_{i=0}^{m_g}\boldsymbol{\gamma}_{l,i}G_{i,[-1\ \tau]}\right)\\ &= \sum_{i=0}^{\max\{2m_q+n_f,m_v,m_g\}+1}\mathbf{s}_{l+1,i}F_i\left(\tau\right) \overset{\text{polynomial truncation}}{\approx} \sum_{i=0}^{m_v}\mathbf{s}_{l+1,i}F_i\left(\tau\right)\end{aligned} \tag{37}$$

and

$$\begin{aligned}\mathbf{p}_{l+1}^e &= \mathbf{p}^e\left(0\right) + \frac{t_N}{2}\int_{-1}^{\tau}\sum_{i=0}^{m_p}\mathbf{s}_{l,i}F_i\left(\tau\right)d\tau = \mathbf{p}^e\left(0\right) + \frac{t_N}{2}\sum_{i=0}^{m_p}\mathbf{s}_{l,i}G_{i,[-1\ \tau]}\\ &= \sum_{i=0}^{m_p+1}\boldsymbol{\rho}_{l+1,i}F_i\left(\tau\right) \overset{\text{polynomial truncation}}{\approx} \sum_{i=0}^{m_p}\boldsymbol{\rho}_{l+1,i}F_i\left(\tau\right)\end{aligned} \tag{38}$$

The last approximations for both velocity and position are due to the polynomial truncations with prescribed maximum degrees, say $m_v$ and $m_p$ as in (34), respectively. As a vector cross product involves 6 scalar multiplications, the velocity computation complexity is roughly proportional to $\mathrm{O}\left(32m_q^2 n_f + 6m_v + m_g\right)$ at each iteration, disregarding the cost in (36). The position computation burden, roughly proportional to $\mathrm{O}\left(m_p\right)$, is usually negligible as compared with attitude and velocity. Similarly, the above iterative process can be repeated until some prescribed maximum iteration number is reached or some convergence criterion is met.

To summarize, the flowchart of the iNavFIter in the Earth frame is presented in Figure 2. Note that the iterative velocity/position computation will not be initiated until the attitude iterative computation has finished. The velocity and position update order could be alternated and the immediate velocity/position result could also be used within the current iteration instead of that at the last iteration. These options, however, would make little difference to the final velocity/position computation accuracy as long as the iterations are sufficient.

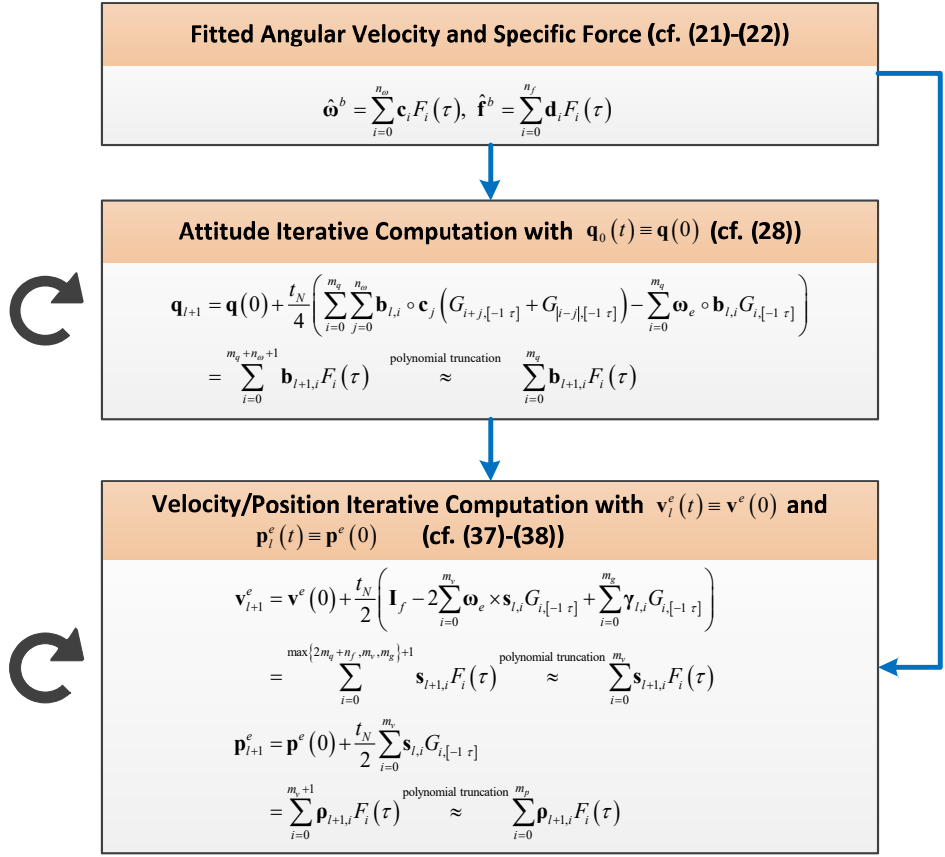

Figure 2. Flowchart of iNavFIter in Earth frame. Arrowed circles mean that two iterative computation processes continue until convergence condition or maximum iteration number is reached.

## IV. Relation to Typical Algorithms

The practical strapdown inertial navigation algorithms typically use two samples of gyroscope/accelerometer incremental measurements for the coning/sculling/scrolling corrections [5, 8, 10], so in this section we briefly compare them to the proposed iNavFIter ( $N = 2$ ). Denote by $T$ the attitude update interval. Table I lists two 2-sample inertial navigation algorithms: one is the typical version [5]; the other is an improved typical version [8]. Readers are referred to [5, 8] and references therein for more details.

As two samples can be used to fit the angular velocity/specific force, the maximum degrees of Chebyshev polynomials are $n_\omega = n_f = 1$. Although the typical algorithms employ the rotation vector for attitude parameterization, it is clear from Table I that the coning correction only involves the cross product between the two raw angular increments. It amounts to the iNavFIter's attitude computation (28) at the 2nd iteration, irrespective of the normalization issue of quaternion.

Both typical algorithms take the first-order approximation of the attitude matrix in integrating the transformed specific force, which corresponds to an one-order truncation of the final quaternion result in (33), namely $m_q = 1$. The typical version uses the zero-order approximation of velocity and position in the velocity update ($m_v = m_p = 0$), and the first-order approximation of velocity in the position update ($m_v = 1$); in contrast, the improved typical version uses the first-order approximation in both the velocity and position updates ($m_v = m_p = 1$). The gravity is unexceptionally taken as a constant within the update interval ($m_g = 0$). All of these amounts to the iNavFIter's velocity/position computation (28) at the 1st iteration.

Another major difference is that the typical 2-sample algorithms can only yield the navigation information at the end of the update interval, while the iNavFIter produces all of attitude/velocity/position information over the whole update interval. For a time interval with more than two samples, the modern-day algorithm designed for $N > 2$ might likely lead to degraded (instead of improved) accuracy due to the simplified rotation vector [27, 51], so the practitioners usually divide the time interval into 2-sample subintervals and consecutively apply the 2-sample algorithm to each subinterval. In principle, the iNavFIter is capable of handling intervals with any number of samples.

Table I. Two 2-sample Inertial Navigation Algorithms in Local-level Navigation Frame

| 2-sample Algorithm | Typical Version [4-6] | Improved Typical Version [8] |
|---|---|---|
| Attitude Update | | $\mathbf{q}_{b(0)}^{b(t)} = \cos\frac{|\boldsymbol{\sigma}_b|}{2} + \frac{\boldsymbol{\sigma}_b}{|\boldsymbol{\sigma}_b|}\sin\frac{|\boldsymbol{\sigma}_b|}{2},\ \boldsymbol{\sigma}_b = \Delta\tilde{\boldsymbol{\theta}}_{t_1} + \Delta\tilde{\boldsymbol{\theta}}_{t_2} + \frac{2}{3}\Delta\tilde{\boldsymbol{\theta}}_{t_1} \times \Delta\tilde{\boldsymbol{\theta}}_{t_2}$ $\mathbf{q}_{n(0)}^{n(t)} = \cos\frac{|\boldsymbol{\sigma}_n|}{2} + \frac{\boldsymbol{\sigma}_n}{|\boldsymbol{\sigma}_n|}\sin\frac{|\boldsymbol{\sigma}_n|}{2},\ \boldsymbol{\sigma}_n = T\boldsymbol{\omega}_{in}^n$ $\mathbf{C}_b^n(t) = \mathbf{C}_{n(0)}^{n(t)}\mathbf{C}_b^n(0)\mathbf{C}_{b(t)}^{b(0)}$ |
| Velocity Update | $\mathbf{v}^n(t) = \mathbf{v}^n(0) + \mathbf{u}$ $-T\left(2\boldsymbol{\omega}_{ie}^n + \boldsymbol{\omega}_{en}^n\right) \times \mathbf{v}^n(0) + T\mathbf{g}^n(0)$ | $\mathbf{u} = \mathbf{C}_b^n(0)\left(\begin{array}{l}\Delta\tilde{\mathbf{v}}_{t_1} + \Delta\tilde{\mathbf{v}}_{t_2} + \frac{1}{2}\left(\Delta\tilde{\boldsymbol{\theta}}_{t_1} + \Delta\tilde{\boldsymbol{\theta}}_{t_2}\right) \times \left(\Delta\tilde{\mathbf{v}}_{t_1} + \Delta\tilde{\mathbf{v}}_{t_2}\right) \\ + \frac{2}{3}\left(\Delta\tilde{\boldsymbol{\theta}}_{t_1} \times \Delta\tilde{\mathbf{v}}_{t_2} + \Delta\tilde{\mathbf{v}}_{t_1} \times \Delta\tilde{\boldsymbol{\theta}}_{t_2}\right)\end{array}\right)$ $\mathbf{v} = \mathbf{C}_{n(0)}^{n(t)}\left\{\begin{array}{l}\mathbf{v}^n(0) + \mathbf{u} - \left(T\mathbf{I}_3 + \frac{T^2}{2}\boldsymbol{\omega}_{in}^n \times\right)\boldsymbol{\omega}_{ie}^n \times \mathbf{v}^n(0) \\ + \left(T\mathbf{I}_3 + \frac{T^2}{2}\boldsymbol{\omega}_{in}^n \times\right)\mathbf{g}^n(0)\end{array}\right\}$ $\mathbf{v}^n(t) = \mathbf{v} - \mathbf{C}_{n(0)}^{n(t)}\left(\frac{T}{2}\mathbf{I}_3 + \frac{T^2}{3}\boldsymbol{\omega}_{in}^n \times\right)\boldsymbol{\omega}_{ie}^n \times \left(\mathbf{v} - \mathbf{v}^n(0)\right)$ |
| Position Update | $\mathbf{r}^n(t) = \frac{T}{2}\left(\mathbf{v}^n(0) + \mathbf{v}^n(t)\right)$ | $I_{\mathbf{u}} = \frac{T}{30}\mathbf{C}_b^n(0)\left(\begin{array}{l}25\Delta\tilde{\mathbf{v}}_{t_1} + 5\Delta\tilde{\mathbf{v}}_{t_2} + 12\Delta\tilde{\boldsymbol{\theta}}_{t_1} \times \Delta\tilde{\mathbf{v}}_{t_1} \\ + 8\Delta\tilde{\boldsymbol{\theta}}_{t_1} \times \Delta\tilde{\mathbf{v}}_{t_2} + 2\Delta\tilde{\mathbf{v}}_{t_1} \times \Delta\tilde{\boldsymbol{\theta}}_{t_2} + 2\Delta\tilde{\boldsymbol{\theta}}_{t_2} \times \Delta\tilde{\mathbf{v}}_{t_2}\end{array}\right)$ $\mathbf{r} = \mathbf{C}_{n(0)}^{n(t)}\left[\begin{array}{l}T\mathbf{v}^n(0) + I_{\mathbf{u}} - \left(\frac{T^2}{3}\mathbf{I}_3 + \frac{T^3}{12}\boldsymbol{\omega}_{in}^n \times\right)\boldsymbol{\omega}_{ie}^n \times \mathbf{v}^n(0) \\ - \left(\frac{T^2}{6}\mathbf{I}_3 + \frac{T^3}{12}\boldsymbol{\omega}_{in}^n \times\right)\boldsymbol{\omega}_{ie}^n \times \mathbf{v}^n(t) + \left(\frac{T^2}{2}\mathbf{I}_3 + \frac{T^3}{6}\boldsymbol{\omega}_{in}^n \times\right)\mathbf{g}^n(0)\end{array}\right]$ $\mathbf{r}^n(t) = \mathbf{r} + \mathbf{C}_{n(0)}^{n(t)}\left(\frac{T}{2}\mathbf{I}_3 + \frac{T^3}{3}\boldsymbol{\omega}_{in}^n \times\right)\boldsymbol{\omega}_{in}^n \times \mathbf{r}$ |
| | $\mathbf{p}(t) = \mathbf{p}(0) + \mathbf{R}_c(0)\mathbf{r}^n(t)$ | |

<div align="center">V. CONVERGENCE PROPERTY AND ERROR ANALYSIS OF iNavFIter</div>

*A. Convergence Property*

*Theorem 1*: Given the true angular velocity function $\boldsymbol{\omega}^b$ over the interval $\begin{bmatrix} 0 & t \end{bmatrix}$, the iterative process as given in (25) converges to the true attitude quaternion function solution to (11) for bounded $t \sup |\boldsymbol{\omega}^b|$.

Proof. For the convenience of convergence analysis, the sequence of quaternion functions in (25) is written as

$$\mathbf{q}_l = \mathbf{q}_0 + \sum_{j=0}^{l-1} (\mathbf{q}_{j+1} - \mathbf{q}_j), \quad l \geq 1 \tag{39}$$

Denote $\alpha = \sup_{\tau \in [0 \ t]} |\boldsymbol{\omega}^b| + \Omega$, where $|\cdot|$ denotes the vector Euclidean norm. Then,

$$|\mathbf{q}_0| = |\mathbf{q}(0)| = 1$$

$$|\mathbf{q}_1 - \mathbf{q}_0| = \left| \frac{1}{2} \int_0^t (\mathbf{q}_0 \circ \boldsymbol{\omega}^b - \boldsymbol{\omega}_e \circ \mathbf{q}_0) d\tau \right| = \frac{1}{2} \left| \int_0^t \left( \begin{bmatrix} \bar{\boldsymbol{\omega}}^b \end{bmatrix} - \begin{bmatrix} \overset{+}{\boldsymbol{\omega}}_e \end{bmatrix} \right) \mathbf{q}_0 \, d\tau \right|$$

$$\leq \frac{1}{2} \int_0^t \left\| \begin{bmatrix} \bar{\boldsymbol{\omega}}^b \end{bmatrix} \right\| + \left\| \begin{bmatrix} \overset{+}{\boldsymbol{\omega}}_e \end{bmatrix} \right\| d\tau \leq \int_0^t (|\boldsymbol{\omega}| + \Omega) d\tau \leq \alpha t$$

$$|\mathbf{q}_2 - \mathbf{q}_1| = \left| \frac{1}{2} \int_0^t \left[ (\mathbf{q}_1 - \mathbf{q}_0) \circ \boldsymbol{\omega}^b - \boldsymbol{\omega}_e \circ (\mathbf{q}_1 - \mathbf{q}_0) \right] d\tau \right| = \frac{1}{2} \left\| \int_0^t \left( \begin{bmatrix} \bar{\boldsymbol{\omega}}^b \end{bmatrix} - \begin{bmatrix} \overset{+}{\boldsymbol{\omega}}_e \end{bmatrix} \right) (\mathbf{q}_1 - \mathbf{q}_0) d\tau \right\| \tag{40}$$

$$\leq \frac{1}{2} \int_0^t \left( \left\| \begin{bmatrix} \bar{\boldsymbol{\omega}}^b \end{bmatrix} \right\| + \left\| \begin{bmatrix} \overset{+}{\boldsymbol{\omega}}_e \end{bmatrix} \right\| \right) |\mathbf{q}_1 - \mathbf{q}_0| d\tau \leq \int_0^t \alpha^2 \tau \, d\tau = \frac{1}{2} \alpha^2 t^2$$

$$\vdots$$

$$|\mathbf{q}_{j+1} - \mathbf{q}_j| = \left| \frac{1}{2} \int_0^t \left[ (\mathbf{q}_j - \mathbf{q}_{j-1}) \circ \boldsymbol{\omega}^b - \boldsymbol{\omega}_e \circ (\mathbf{q}_j - \mathbf{q}_{j-1}) \right] d\tau \right| \leq \frac{1}{(j+1)!} (\alpha t)^{j+1}$$

where $\|\cdot\|$ denotes the matrix Frobenius norm and $\left\| \begin{bmatrix} \bar{\mathbf{q}} \end{bmatrix} \right\| = \left\| \begin{bmatrix} \overset{+}{\mathbf{q}} \end{bmatrix} \right\| = 2 |\mathbf{q}|$ for any quaternion according to the definition of the quaternion multiplication matrix in (6). Note that the sum of the bounds on the right side converges, namely, $1 + \alpha t + \frac{1}{2} \alpha^2 t^2 + \cdots + \frac{1}{(j+1)!} (\alpha t)^{j+1} + \cdots = e^{\alpha t}$. According to the Weierstrass M-Test [30], the iterative process (25) converges uniformly and absolutely on the interval $\begin{bmatrix} 0 & t \end{bmatrix}$. Finally, it can be readily checked that the limit of the quaternion function

sequence satisfies (11), i.e., $\dot{\mathbf{q}}_\infty = \left(\mathbf{q}_\infty \circ \boldsymbol{\omega}^b - \boldsymbol{\omega}_e \circ \mathbf{q}_\infty\right)\big/2$. It means that $\mathbf{q}_\infty$ is the solution to the quaternion kinematic equation (11), that it to say, the iterative process (25) converges to the true attitude quaternion function.

■

Newly define a combination vector of velocity and position as $\mathbf{x} = \begin{bmatrix} \mathbf{v}^{eT} & \mathbf{p}^{eT} \end{bmatrix}^T$. Then the iterative velocity and position computation in (31) and (32) can be collectively written as

$$\mathbf{x}_{l+1} = \mathbf{x}(0) + \mathbf{W}\int_0^t \mathbf{x}_l \, dt + \mathbf{L}\left(\int_0^t \mathbf{C}_b^e(\mathbf{q})\mathbf{f}^b dt + \int_0^t \mathbf{g}^e(\mathbf{H}\mathbf{x}_l) \, dt\right) \tag{41}$$

where the constant matrices $\mathbf{W} = \begin{bmatrix} -2\boldsymbol{\omega}_e \times & \mathbf{0}_{3\times3} \\ \mathbf{I}_3 & \mathbf{0}_{3\times3} \end{bmatrix}$, $\mathbf{L} = \begin{bmatrix} \mathbf{I}_3 \\ \mathbf{0}_{3\times3} \end{bmatrix}$, and $\mathbf{H} = \begin{bmatrix} \mathbf{0}_{3\times3} & \mathbf{I}_3 \end{bmatrix}$. Denote

$\beta = \|\mathbf{W}\|\|\mathbf{x}_0\| + \|\mathbf{L}\|\left(\sup\limits_{\tau \in [0 \quad t]}\left|\mathbf{f}^b\right| + \sup\limits_{\tau \in [0 \quad t]}\left|\mathbf{g}^e\right|\right)$ and $\gamma = \|\mathbf{W}\| + \|\mathbf{L}\|\sup\limits_{\tau \in [0 \quad t]}\left\|\dfrac{\partial \mathbf{g}^e}{\partial (\mathbf{H}\mathbf{x})}\right\|\|\mathbf{H}\|$.

*Theorem 2*: Given the true specific force function $\mathbf{f}^b$ over the interval $\begin{bmatrix} 0 & t \end{bmatrix}$, the iterative process as given in (31)-(32) converges to the true velocity/position function solution to (12)-(13), respectively, for bounded $\beta$ and $t\gamma$.

Proof. The sequence of velocity/position vector functions is rewritten as

$$\mathbf{x}_l = \mathbf{x}_0 + \sum_{j=0}^{l-1}\left(\mathbf{x}_{j+1} - \mathbf{x}_j\right), \quad l \geq 1 \tag{42}$$

Then,

$$\left|\mathbf{x}_0\right| = \left|\begin{bmatrix} \mathbf{v}^{eT}(0) & \mathbf{p}^{eT}(0) \end{bmatrix}^T\right|$$

$$\left|\mathbf{x}_1 - \mathbf{x}_0\right| = \left|\mathbf{W}\int_0^t \mathbf{x}_0 d\tau + \mathbf{L}\left(\int_0^t \mathbf{C}_b^e(\mathbf{q})\mathbf{f}^b d\tau + \int_0^t \mathbf{g}^e(\mathbf{H}\mathbf{x}_0) d\tau\right)\right| \leq \|\mathbf{W}\|\int_0^t |\mathbf{x}_0| d\tau + \|\mathbf{L}\|\left(\int_0^t \left|\mathbf{f}^b\right| d\tau + \int_0^t \left|\mathbf{g}^e(\mathbf{H}\mathbf{x}_0)\right| d\tau\right)$$

$$\leq t\left[\|\mathbf{W}\|\|\mathbf{x}_0\| + \|\mathbf{L}\|\left(\sup\left|\mathbf{f}^b\right| + \left|\mathbf{g}^e(\mathbf{H}\mathbf{x}_0)\right|\right)\right] \leq t\beta$$

$$\left|\mathbf{x}_2 - \mathbf{x}_1\right| = \left|\mathbf{W}\int_0^t \mathbf{x}_1 - \mathbf{x}_0 d\tau + \mathbf{L}\int_0^t \mathbf{g}^e(\mathbf{H}\mathbf{x}_1) - \mathbf{g}^e(\mathbf{H}\mathbf{x}_0) d\tau\right| \leq \|\mathbf{W}\|\int_0^t |\mathbf{x}_1 - \mathbf{x}_0| d\tau + \|\mathbf{L}\|\int_0^t \left|\mathbf{g}^e(\mathbf{H}\mathbf{x}_1) - \mathbf{g}^e(\mathbf{H}\mathbf{x}_0)\right| d\tau \tag{43}$$

$$\approx \|\mathbf{W}\|\int_0^t |\mathbf{x}_1 - \mathbf{x}_0| d\tau + \|\mathbf{L}\|\int_0^t \left|\frac{\partial \mathbf{g}^e}{\partial (\mathbf{H}\mathbf{x}_0)}\mathbf{H}(\mathbf{x}_1 - \mathbf{x}_0)\right| d\tau \leq \left(\|\mathbf{W}\| + \|\mathbf{L}\|\sup\left\|\frac{\partial \mathbf{g}^e}{\partial (\mathbf{H}\mathbf{x}_0)}\right\|\|\mathbf{H}\|\right)\int_0^t |\mathbf{x}_1 - \mathbf{x}_0| d\tau \leq \frac{t^2\gamma}{2}\beta$$

$$\vdots$$

$$\left|\mathbf{x}_{j+1} - \mathbf{x}_j\right| = \left|\mathbf{W}\int_0^t \mathbf{x}_j - \mathbf{x}_{j-1} d\tau + \mathbf{L}\int_0^t \mathbf{g}^e(\mathbf{H}\mathbf{x}_j) - \mathbf{g}^e(\mathbf{H}\mathbf{x}_{j-1}) d\tau\right| \leq \frac{t^{j+1}\gamma^j}{(j+1)!}\beta$$

Note that the sum of the bounds on the right side $|\mathbf{x}_0| + t\beta + \dfrac{(t\gamma)^2}{2}\dfrac{\beta}{\gamma} + \cdots + \dfrac{(t\gamma)^{j+1}}{(j+1)!}\dfrac{\beta}{\gamma} + \cdots = |\mathbf{x}_0| + \dfrac{\beta}{\gamma}\left(e^{t\gamma} - 1\right)$. According to the Weierstrass M-Test again, the iterative processes (31) and (32) converge uniformly and absolutely on the interval $\begin{bmatrix} 0 & t \end{bmatrix}$. Finally, it can be readily checked that the limit of the function sequence satisfies (12)-(13), i.e., $\dot{\mathbf{x}}_\infty = \mathbf{W}\mathbf{x}_\infty + \mathbf{L}\mathbf{C}_b^e(\mathbf{q})\mathbf{f}^b + \mathbf{g}^e(\mathbf{H}\mathbf{x}_\infty)$. It

means that $\mathbf{x}_\infty$ is the solution to the velocity/position kinematic equations, that it to say, the iterative processes (31) and (32) converge to the true velocity and position function, respectively.

∎

### B. Error Analysis

Theorems 1-2 in the last subsection guarantee that the iterative iNavFIter would yield the true navigation parameters for error-free inertial measurements. In general, however, the algorithm has three kinds of error sources, namely, the sensor error ($\delta\boldsymbol{\omega}^b$ and $\delta\mathbf{f}^b$), the initial iteration error ($\delta\mathbf{q}_0$, $\delta\mathbf{v}_0^e$ and $\delta\mathbf{p}_0^e$) and the truncation error at the $l$-th iteration ($\delta\mathbf{q}_l^t$, $\delta\mathbf{v}_l^{et}$ and $\delta\mathbf{p}_l^{et}$).

According to (25), the additive attitude quaternion error up to the first order propagates as

$$
\begin{aligned}
|\delta\mathbf{q}_1| &= \left| \frac{1}{2}\int_0^t \left( \left[ \bar{\boldsymbol{\omega}}^b \right] \delta\mathbf{q}_0 + \left[ \delta\bar{\boldsymbol{\omega}}^b \right] \mathbf{q}_0 - \left[ \bar{\boldsymbol{\omega}}_e \right] \delta\mathbf{q}_0 \right) dt + \delta\mathbf{q}_1^t \right| \\
&\leq \frac{1}{2}\left( \int_0^t \left( \left\| \left[ \bar{\boldsymbol{\omega}}^b \right] \right\| + \left\| \left[ \bar{\boldsymbol{\omega}}_e \right] \right\| \right) |\delta\mathbf{q}_0| \, dt + \int_0^t \left\| \left[ \delta\bar{\boldsymbol{\omega}}^b \right] \right\| |\mathbf{q}_0| \, dt \right) + \left| \delta\mathbf{q}_1^t \right| \\
&\leq t\left( \sup|\boldsymbol{\omega}^b| + \Omega \right) \sup|\delta\mathbf{q}_0| + t\sup|\delta\boldsymbol{\omega}^b| \sup|\mathbf{q}_0| + \left| \delta\mathbf{q}_1^t \right|
\end{aligned}
\tag{44}
$$

and then following the development of (24)-(25) in [30], we have

$$
\begin{aligned}
|\delta\mathbf{q}_2| &\leq \frac{1}{2}\left( \int_0^t \left( \left\| \left[ \bar{\boldsymbol{\omega}}^b \right] \right\| + \left\| \left[ \bar{\boldsymbol{\omega}}_e \right] \right\| \right) |\delta\mathbf{q}_1| \, dt + \int_0^t \left\| \left[ \delta\bar{\boldsymbol{\omega}}^b \right] \right\| |\mathbf{q}_1| \, dt \right) + \left| \delta\mathbf{q}_2^t \right| \\
&\leq \left( \sup|\boldsymbol{\omega}^b| + \Omega \right) \int_0^t |\delta\mathbf{q}_1| \, dt + t\sup|\delta\boldsymbol{\omega}^b| \sup|\mathbf{q}_1| + \left| \delta\mathbf{q}_2^t \right| \\
&\leq \left( \sup|\boldsymbol{\omega}^b| + \Omega \right) \int_0^t \left[ t\left( \sup|\boldsymbol{\omega}^b| + \Omega \right) \sup|\delta\mathbf{q}_0| + t\sup|\delta\boldsymbol{\omega}^b| \sup|\mathbf{q}_0| + \left| \delta\mathbf{q}_1^t \right| \right] dt + t\sup|\delta\boldsymbol{\omega}^b| \sup|\mathbf{q}_1| + \left| \delta\mathbf{q}_2^t \right| \\
&= \frac{t^2}{2}\left( \sup|\boldsymbol{\omega}^b| + \Omega \right)^2 \sup|\delta\mathbf{q}_0| + \frac{t^2}{2}\left( \sup|\boldsymbol{\omega}^b| + \Omega \right) \sup|\delta\boldsymbol{\omega}^b| \sup|\mathbf{q}_0| + t\sup|\delta\boldsymbol{\omega}^b| \sup|\mathbf{q}_1| + t\left( \sup|\boldsymbol{\omega}^b| + \Omega \right) \sup|\delta\mathbf{q}_1^t| + \left| \delta\mathbf{q}_2^t \right|
\end{aligned}
\tag{45}
$$

$$
\cdots
$$

$$
\begin{aligned}
|\delta\mathbf{q}_l| &\leq \frac{1}{2}\left( \int_0^t \left( \left\| \left[ \bar{\boldsymbol{\omega}}^b \right] \right\| + \left\| \left[ \bar{\boldsymbol{\omega}}_e \right] \right\| \right) |\delta\mathbf{q}_{l-1}| \, dt + \int_0^t \left\| \left[ \delta\bar{\boldsymbol{\omega}}^b \right] \right\| |\mathbf{q}_{l-1}| \, dt \right) + \left| \delta\mathbf{q}_l^t \right| \\
&\leq \frac{t^l\left( \sup|\boldsymbol{\omega}^b| + \Omega \right)^l}{l!}\sup|\delta\mathbf{q}_0| + t\sup|\delta\boldsymbol{\omega}^b| \sum_{k=0}^{l-1} \frac{t^k\left( \sup|\boldsymbol{\omega}^b| + \Omega \right)^k}{(k+1)!}\sup|\mathbf{q}_{l-1-k}| + \sum_{k=0}^{l-1} \frac{t^k\left( \sup|\boldsymbol{\omega}^b| + \Omega \right)^k}{k!}\sup|\delta\mathbf{q}_{l-k}^t|
\end{aligned}
$$

Regarding (45), the first term is owed to the initial quaternion error and quickly vanishes for large iterations for any bounded $t\left( \sup|\boldsymbol{\omega}^b| + \Omega \right)$, which means the attitude computation converges regardless of the initial quaternion function. The third term is owed to the polynomial truncation at each iteration, in which the weights of the early iterations are much smaller than those of later iterations, and thus can be approximated by the last truncation error, i.e.,

$$
\sum_{k=0}^{l-1} \frac{t^k\left( \sup|\boldsymbol{\omega}^b| + \Omega \right)^k}{k!}\sup|\delta\mathbf{q}_{l-k}^t| \approx \sum_{k=0}^{l-1} \frac{t^k\left( \sup|\boldsymbol{\omega}^b| + \Omega \right)^k}{k!}\left| \mathbf{b}_{l-k, m_q+1} \right| \approx \left| \mathbf{b}_{l, m_q+1} \right|.
\tag{46}
$$

The second term depends on the angular velocity error and the norm of the quaternion estimate, and it can be approximated by

$$t\sup\left|\delta\boldsymbol{\omega}^b\right|\sum_{k=0}^{l-1}\frac{t^k\left(\sup\left|\boldsymbol{\omega}^b\right|+\Omega\right)^k}{(k+1)!}\sup\left|\mathbf{q}_{l-1-k}\right|\approx t\sup\left|\delta\boldsymbol{\omega}^b\right|\sum_{k=0}^{l-1}\frac{t^k\left(\sup\left|\boldsymbol{\omega}^b\right|+\Omega\right)^k}{(k+1)!}\approx t\sup\left|\delta\boldsymbol{\omega}^b\right| \tag{47}$$

as $\sup\left|\mathbf{q}\right|\approx1$. Therefore, for a large number of iterations, the attitude quaternion error is approximately bounded by

$$\sup\left|\delta\mathbf{q}_l\right|\approx t\sup\left|\delta\boldsymbol{\omega}^b\right|+\left|\mathbf{b}_{l,m_q+1}\right|. \tag{48}$$

It means that the error of iterative attitude computation (28) is generally dominated by the angular velocity error and the last truncation error as well. An ideal case is when $\delta\boldsymbol{\omega}^b=0$, for which higher order of truncation means higher accuracy.

Using (41), the first-order error of velocity/position propagates as

$$\begin{aligned}
\left|\delta\mathbf{x}_1\right| &= \left|\mathbf{W}\int_0^t\delta\mathbf{x}_0dt+\mathbf{L}\left[\int_0^t\left(\delta\mathbf{q}\circ\mathbf{f}^b\circ\mathbf{q}^*+\mathbf{q}\circ\delta\mathbf{f}^b\circ\mathbf{q}^*+\mathbf{q}\circ\mathbf{f}^b\circ\delta\mathbf{q}^*\right)dt+\int_0^t\frac{\partial\mathbf{g}^e}{\partial(\mathbf{Hx}_0)}\mathbf{H}\delta\mathbf{x}_0dt\right]+\delta\mathbf{x}_1'\right| \\
&\le \left(\|\mathbf{W}\|+\|\mathbf{L}\|\sup\left\|\frac{\partial\mathbf{g}^e}{\partial(\mathbf{Hx}_0)}\right\|\|\mathbf{H}\|\right)\int_0^t\left|\delta\mathbf{x}_0\right|dt+\|\mathbf{L}\|\left|\int_0^t\left(\delta\mathbf{q}\circ\mathbf{f}^b\circ\mathbf{q}^*+\mathbf{q}\circ\delta\mathbf{f}^b\circ\mathbf{q}^*+\mathbf{q}\circ\mathbf{f}^b\circ\delta\mathbf{q}^*\right)dt\right|+\left|\delta\mathbf{x}_1'\right| \\
&\le t\left(\|\mathbf{W}\|+\|\mathbf{L}\|\sup\left\|\frac{\partial\mathbf{g}^e}{\partial(\mathbf{Hx}_0)}\right\|\|\mathbf{H}\|\right)\sup\left|\delta\mathbf{x}_0\right|+t\|\mathbf{L}\|\left(8\sup\left|\mathbf{f}^b\right|\sup\left|\delta\mathbf{q}\right|+\sup\left|\delta\mathbf{f}^b\right|\right)+\left|\delta\mathbf{x}_1'\right|
\end{aligned} \tag{49}$$

where the relationship $\delta\mathbf{q}^*=-\mathbf{q}^*\circ\delta\mathbf{q}\circ\mathbf{q}^*$ is used. Then we have

$$\begin{aligned}
\left|\delta\mathbf{x}_2\right| &\le \left(\|\mathbf{W}\|+\|\mathbf{L}\|\sup\left\|\frac{\partial\mathbf{g}^e}{\partial(\mathbf{Hx}_1)}\right\|\|\mathbf{H}\|\right)\int_0^t\left|\delta\mathbf{x}_1\right|dt+\mathbf{L}\left|\int_0^t\left(\delta\mathbf{q}\circ\mathbf{f}^b\circ\mathbf{q}^*+\mathbf{q}\circ\delta\mathbf{f}^b\circ\mathbf{q}^*+\mathbf{q}\circ\mathbf{f}^b\circ\delta\mathbf{q}^*\right)dt\right|+\left|\delta\mathbf{x}_2'\right| \\
&\le \left(\|\mathbf{W}\|+\|\mathbf{L}\|\sup\left\|\frac{\partial\mathbf{g}^e}{\partial(\mathbf{Hx}_1)}\right\|\|\mathbf{H}\|\right)\left\{\frac{t^2}{2}\left(\|\mathbf{W}\|+\|\mathbf{L}\|\sup\left\|\frac{\partial\mathbf{g}^e}{\partial(\mathbf{Hx}_0)}\right\|\|\mathbf{H}\|\right)\sup\left|\delta\mathbf{x}_0\right|+\frac{t^2}{2}\|\mathbf{L}\|\left(8\sup\left|\mathbf{f}^b\right|\sup\left|\delta\mathbf{q}\right|+\sup\left|\delta\mathbf{f}^b\right|\right)+t\left|\delta\mathbf{x}_1'\right|\right\} \\
&+t\|\mathbf{L}\|\left(8\sup\left|\mathbf{f}^b\right|\sup\left|\delta\mathbf{q}\right|+\sup\left|\delta\mathbf{f}^b\right|\right)+\left|\delta\mathbf{x}_2'\right| \\
&= \frac{t^2}{2}\left(\|\mathbf{W}\|+\|\mathbf{L}\|\sup\left\|\frac{\partial\mathbf{g}^e}{\partial(\mathbf{Hx}_1)}\right\|\|\mathbf{H}\|\right)\left(\|\mathbf{W}\|+\|\mathbf{L}\|\sup\left\|\frac{\partial\mathbf{g}^e}{\partial(\mathbf{Hx}_0)}\right\|\|\mathbf{H}\|\right)\sup\left|\delta\mathbf{x}_0\right| \\
&+\left[\frac{t^2}{2}\left(\|\mathbf{W}\|+\|\mathbf{L}\|\sup\left\|\frac{\partial\mathbf{g}^e}{\partial(\mathbf{Hx}_1)}\right\|\|\mathbf{H}\|\right)+t\right]\|\mathbf{L}\|\left(8\sup\left|\mathbf{f}^b\right|\sup\left|\delta\mathbf{q}\right|+\sup\left|\delta\mathbf{f}^b\right|\right)+t\left(\|\mathbf{W}\|+\|\mathbf{L}\|\sup\left\|\frac{\partial\mathbf{g}^e}{\partial(\mathbf{Hx}_1)}\right\|\|\mathbf{H}\|\right)\left|\delta\mathbf{x}_1'\right|+\left|\delta\mathbf{x}_2'\right|
\end{aligned}$$

$$\tag{50}$$

$$\begin{aligned}
\left|\delta\mathbf{x}_l\right| &\le \sup\left|\delta\mathbf{x}_0\right|\frac{1}{l!}\prod_{i=0}^{l-1}t\left(\|\mathbf{W}\|+\|\mathbf{L}\|\sup\left\|\frac{\partial\mathbf{g}^e}{\partial(\mathbf{Hx}_i)}\right\|\|\mathbf{H}\|\right) \\
&+\|\mathbf{L}\|\left(8\sup\left|\mathbf{f}^b\right|\sup\left|\delta\mathbf{q}\right|+\sup\left|\delta\mathbf{f}^b\right|\right)\sum_{i=0}^{l-1}\frac{t^{i+1}}{(i+1)!}\left(\|\mathbf{W}\|+\|\mathbf{L}\|\sup\left\|\frac{\partial\mathbf{g}^e}{\partial(\mathbf{Hx}_i)}\right\|\|\mathbf{H}\|\right)_i \\
&+\sum_{i=0}^{l-1}\frac{t^i}{i!}\left(\|\mathbf{W}\|+\|\mathbf{L}\|\sup\left\|\frac{\partial\mathbf{g}^e}{\partial(\mathbf{Hx}_i)}\right\|\|\mathbf{H}\|\right)_i\left|\delta\mathbf{x}_{l-i}'\right|
\end{aligned} \tag{51}$$

Note that $\left( \|\mathbf{W}\| + \|\mathbf{L}\| \sup \left\| \dfrac{\partial \mathbf{g}^e}{\partial (\mathbf{H}\mathbf{x}_i)} \right\| \|\mathbf{H}\| \right)_i$ is 1 for $i=0$ and otherwise normal as it is. Among the above error components, the

first term is owed to the initial velocity/position error and quickly vanishes for large iterations for any bounded

$t \left( \|\mathbf{W}\| + \|\mathbf{L}\| \sup \left\| \dfrac{\partial \mathbf{g}^e}{\partial (\mathbf{H}\mathbf{x}_i)} \right\| \|\mathbf{H}\| \right)$. It means the iNavFIter converges regardless of the initial velocity/position function. The third

term is owed to the polynomial truncation at each iteration, in which the weights of the early iterations are much smaller than those of later iterations, and thus can be approximated by the last truncation error, i.e.,

$$\sum_{i=0}^{l-1} \frac{t^i}{i!} \left( \|\mathbf{W}\| + \|\mathbf{L}\| \sup \left\| \frac{\partial \mathbf{g}^e}{\partial (\mathbf{H}\mathbf{x}_i)} \right\| \|\mathbf{H}\| \right)_i \left| \delta \mathbf{x}_{l-i}^i \right| \approx \sum_{i=0}^{l-1} \frac{t^i}{i!} \left( \|\mathbf{W}\| + \|\mathbf{L}\| \sup \left\| \frac{\partial \mathbf{g}^e}{\partial (\mathbf{H}\mathbf{x}_i)} \right\| \|\mathbf{H}\| \right)_i \sqrt{\mathbf{s}_{l-i,m_v+1}^2 + \boldsymbol{\rho}_{l-i,m_p+1}^2} \approx \sqrt{\mathbf{s}_{l,m_v+1}^2 + \boldsymbol{\rho}_{l,m_p+1}^2}. \qquad (52)$$

The second term depends on the quaternion error and the specific force error, and with (48) it can be approximated by

$$\begin{aligned} &\|\mathbf{L}\| \left( 8 \sup \left| \mathbf{f}^b \right| \sup \left| \delta \mathbf{q} \right| + \sup \left| \delta \mathbf{f}^b \right| \right) \sum_{i=0}^{l-1} \frac{t^{i+1}}{(i+1)!} \left( \|\mathbf{W}\| + \|\mathbf{L}\| \sup \left\| \frac{\partial \mathbf{g}^e}{\partial (\mathbf{H}\mathbf{x}_i)} \right\| \|\mathbf{H}\| \right)_i \\ &\approx t \|\mathbf{L}\| \left( 8 \sup \left| \mathbf{f}^b \right| \sup \left| \delta \mathbf{q} \right| + \sup \left| \delta \mathbf{f}^b \right| \right) \approx t \|\mathbf{L}\| \left[ 8 \sup \left| \mathbf{f}^b \right| \left( t \sup \left| \delta \boldsymbol{\omega}^b \right| + \left| \mathbf{b}_{m_q+1} \right| \right) + \sup \left| \delta \mathbf{f}^b \right| \right] \end{aligned} \qquad (53)$$

Therefore, for a large number of iterations, the velocity/position error of the fast iNavFIter is approximately bounded by

$$\sup \left| \delta \mathbf{x}_l \right| \approx t \|\mathbf{L}\| \left[ 8 \sup \left| \mathbf{f}^b \right| \left( t \sup \left| \delta \boldsymbol{\omega}^b \right| + \left| \mathbf{b}_{m_q+1} \right| \right) + \sup \left| \delta \mathbf{f}^b \right| \right] + \sqrt{\mathbf{s}_{l,m_v+1}^2 + \boldsymbol{\rho}_{l,m_p+1}^2} \qquad (54)$$

This indicates that the fast iNavFIter's velocity/position error is generally dominated by the angular velocity error, the specific force error and the last truncation error as well. Ideally, when $\delta \boldsymbol{\omega}^b = 0$ and $\delta \mathbf{f}^b = 0$, higher order of truncation means higher accuracy.

Regarding the upper bounds of the attitude errors as in (48) and the velocity/position errors (54), it is the truncation parts, namely

$\left| \mathbf{b}_{l,m_q+1} \right|$ and $\sqrt{\mathbf{s}_{l,m_v+1}^2 + \boldsymbol{\rho}_{l,m_p+1}^2}$, that the iNavFIter algorithm tries to handle as best as possible. How much accuracy improvement it can achieve depends on the relative proportion of the motion-incurred non-commutativity errors with respect to the remaining sensor-error parts.

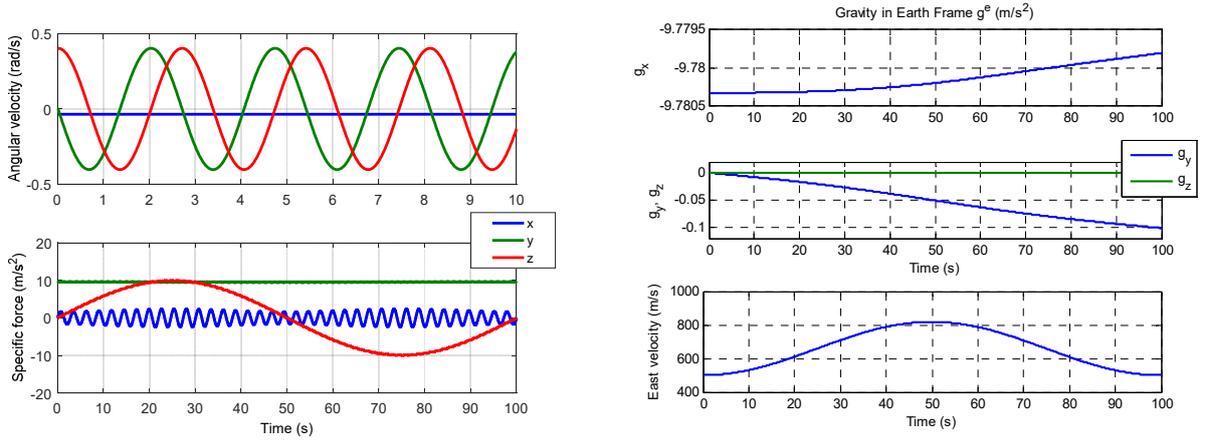

Figure 3. Simulated profile of body angular velocity and specific force in x-y-z axes of body frame (left column), east velocity (lower-right) and gravity (upper-right) in Earth frame.

Table II. Simulation Parameters of Flight Test Trajectory

| Parameter | Value | Unit |
|-----------|-------|------|
| $a$ | 10 | m/s$^2$ |
| $w$ | $0.02\pi$ | rad/s |
| $v_0$ | 500 | m/s |
| $\zeta$ | $0.74\pi$ | rad/s |
| $\alpha$ | 10 | deg |

## VI. NUMERICAL RESULTS

In this section, fight test datasets around the globe are simulated by following our previous work [8], to which the motivation is not for scenario reality but for trajectories with analytical ground truths for the sake of accuracy evaluation. Note that the simulation data is generated in the local-level frame and then transformed into the Earth frame, mainly due to the convenient gravity model in the former frame. Specifically, the popular Somigliana gravity model of WGS-84 is used (see Appendix for details).

Suppose a vehicle carrying a strapdown inertial navigation system flies at the ground velocity $\dot{\mathbf{v}}^n = \begin{bmatrix} 0 & 0 & a\sin(wt) \end{bmatrix}^T$ with an initial east speed $v_0$, where $a$ and $w$ are respectively the magnitude and angular frequency of the velocity rate. The initial position is set to zero longitude, zero latitude and zero height. The body attitude is assumed to undergo a classical coning motion described by the attitude quaternion [4]

$$\mathbf{q}_n^b = \cos(\alpha/2) + \sin(\alpha/2)\begin{bmatrix} 0 & \cos(\zeta t) & \sin(\zeta t) \end{bmatrix}^T \tag{55}$$

where $\zeta$ is the coning frequency and $\alpha$ is the coning angle. Then, the true position and velocity are readily given by

$$\mathbf{v}^n = \mathbf{v}^n(0) + \int_0^t \dot{\mathbf{v}}^n dt = \begin{bmatrix} 0 & 0 & v_0 - (a\cos(wt) - a)/w \end{bmatrix}^T$$

$$\mathbf{p}^n = \int_0^t \mathbf{R}_c \mathbf{v}^n dt = \begin{bmatrix} v_0 t - (a\sin(wt) - awt)/w^2 & 0 & 0 \end{bmatrix}^T / R_E$$

(56)

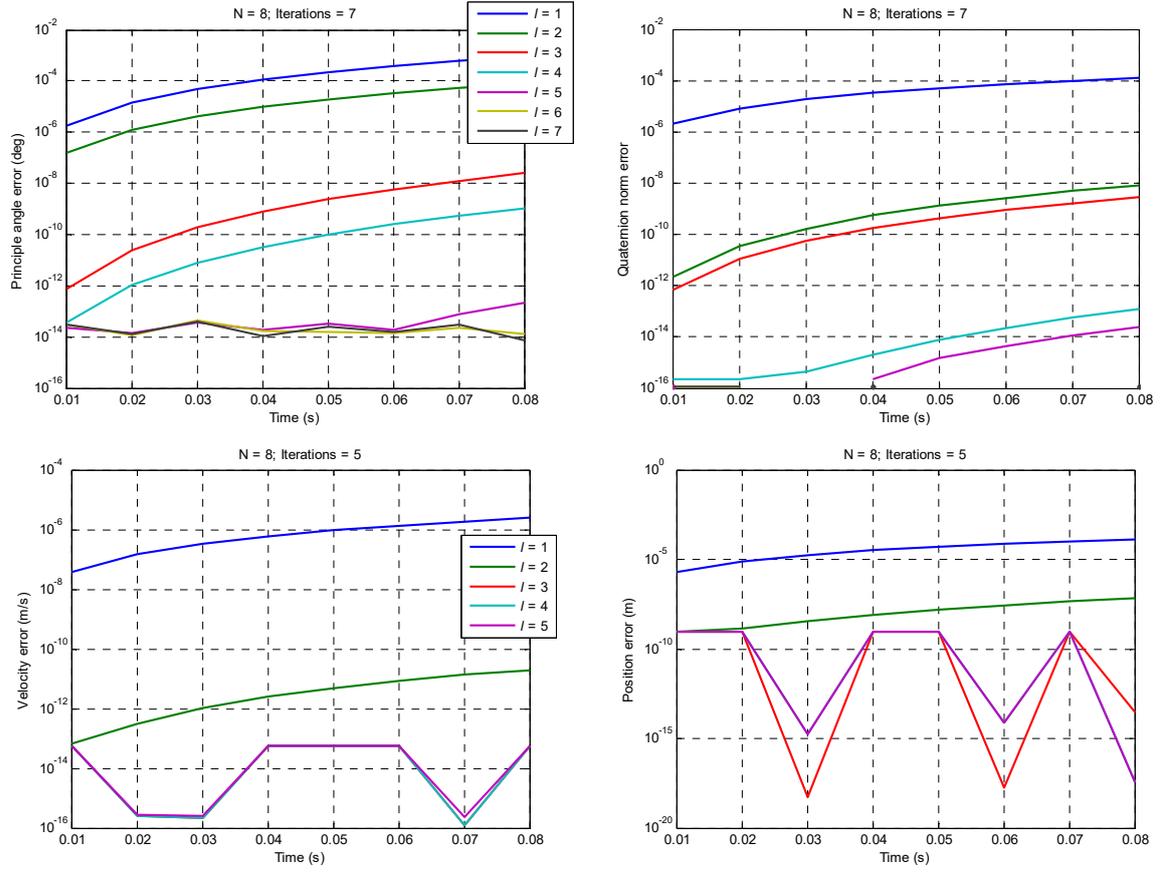

Figure 4. Attitude principal angle error (upper-left), attitude quaternion norm error (upper-right), magnitudes of Earth-frame velocity error (lower-left) and position error (lower-right) in the first update interval.

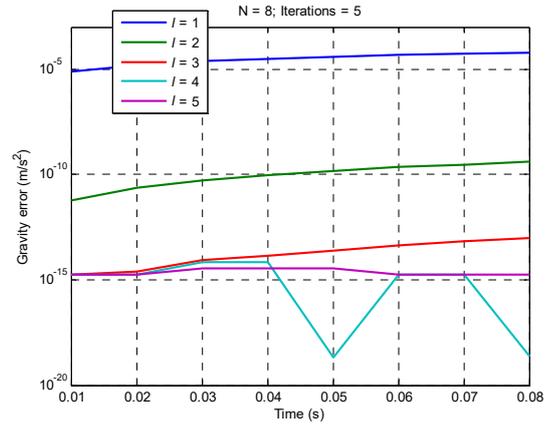

Figure 5. Magnitude of Earth-frame gravity error during velocity/position iterative computation in the first update interval.

From the navigation equations in the local-level frame (1)-(3), the angular velocity and the specific force can be analytically

derived as

$$\boldsymbol{\omega}_{ib}^{b} = \mathbf{q}_{n}^{b*} \circ \left( 2\dot{\mathbf{q}}_{n}^{b} + \boldsymbol{\omega}_{in}^{n} \circ \mathbf{q}_{n}^{b} \right)$$
$$\mathbf{f}^{b} = \mathbf{C}_{n}^{b} \left( \dot{\mathbf{v}}^{n} + \left( 2\boldsymbol{\omega}_{ie}^{n} + \boldsymbol{\omega}_{en}^{n} \right) \times \mathbf{v}^{n} - \mathbf{g}^{n} \right) \tag{57}$$

whose integrals produce the angular increment and the velocity increment to be used as the true gyroscope and accelerometer outputs, respectively. The integrals could be analytically accomplished by the help of the Matlab Symbolic Toolbox. The specific values of the above simulation parameters are listed in Table II. The gyroscope and accelerometer sampling rate is 100 Hz. Figure 3 plots the profile of the body angular velocity for the first 10 seconds, and specific force, east velocity and gravity for the first 100 seconds.

For the iNavFIter algorithm, $N = 8$ samples of angular and velocity increments are used for fitting the angular velocity and specific force, respectively, and the orders of fitted Chebyshev polynomials are set to $n_\omega = n_f = N - 1$. The truncation orders of attitude, velocity and position are uniformly set to $m_q = m_v = m_p = N + 1$. The maximum Chebyshev polynomial degree and the

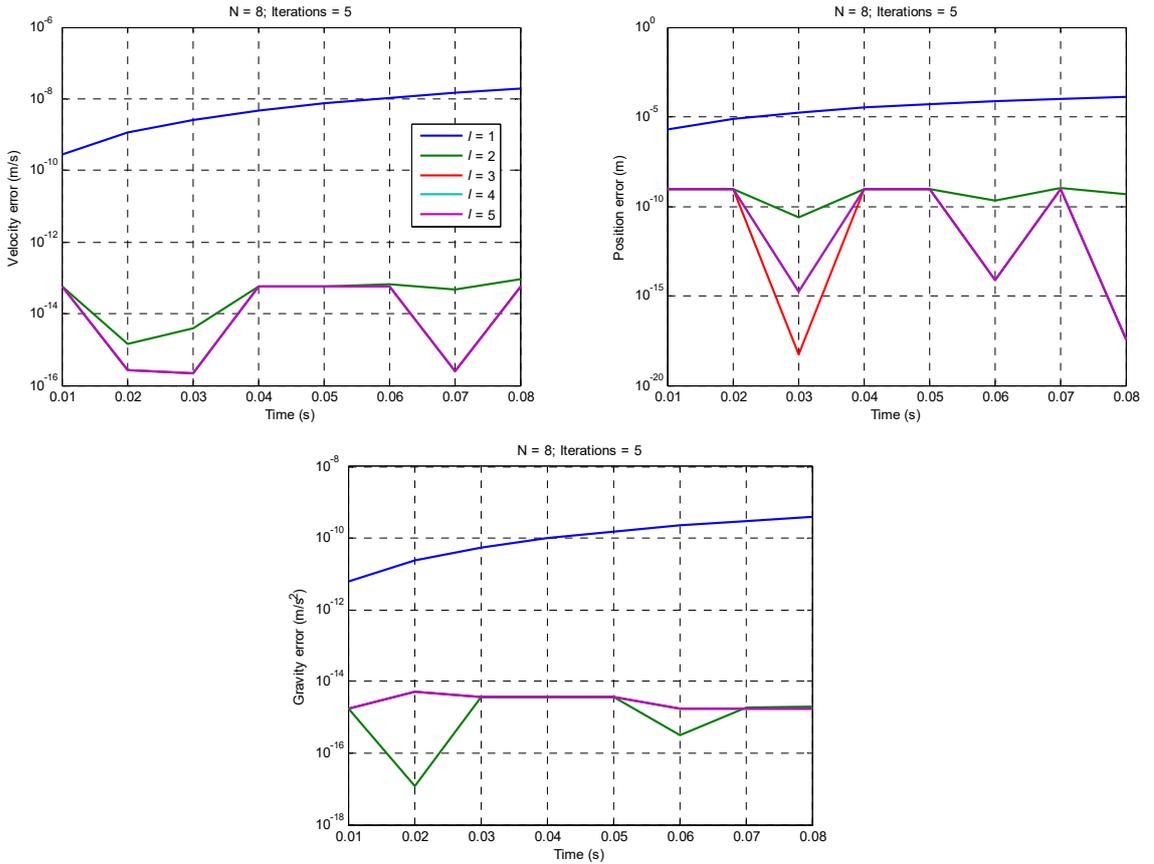

Figure 6. Magnitudes of Earth-frame velocity error (upper-left), position error (upper-right) and gravity error (lower center) in the first update interval, as a result of velocity/position update order change and immediate result usage.

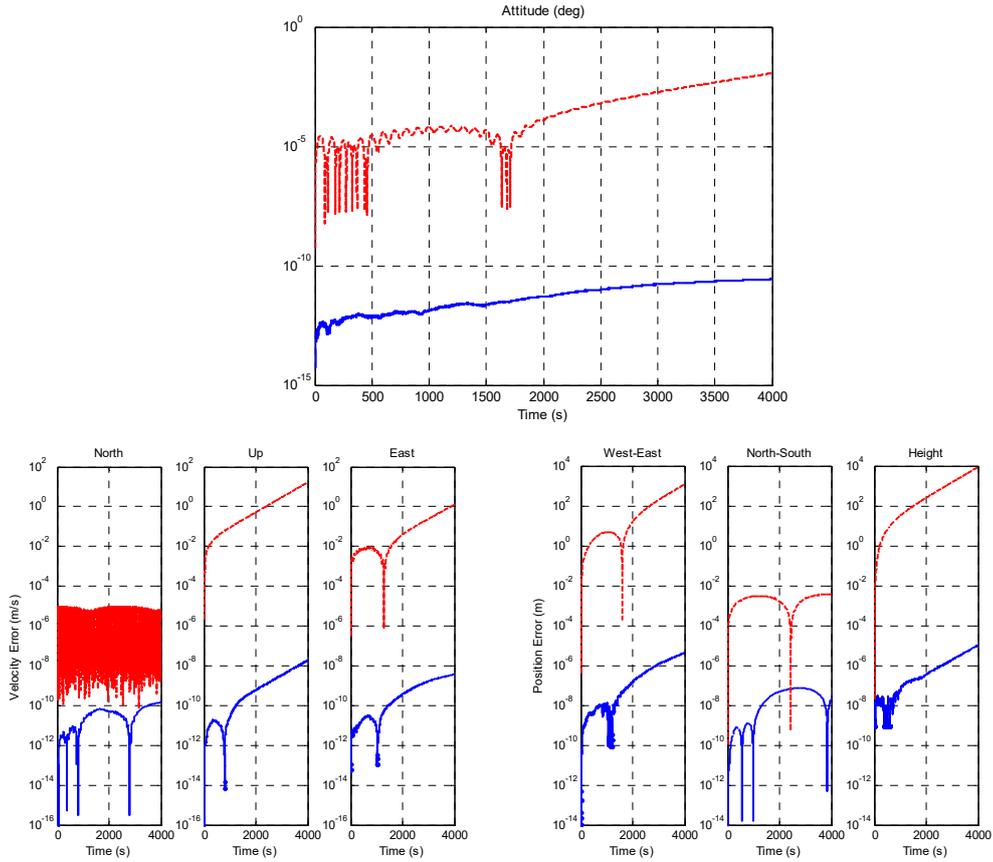

Figure 7. Navigation computation error of iNavFIter algorithm (blue solid line) and typical 2-sample algorithm (red dashed line), for varying-speed coning flight.

number of summation terms for gravity approximation is equally set to $m_g = P = 5$. The maximum iteration numbers of all iterative processes are uniformly set to $N+1$ and the RMS of the discrepancy of the polynomial coefficients between successive iterations is adopted as the convergence criterion and set to $10^{-16}$. Figure 4 presents the attitude principal angle error, the attitude quaternion norm error, the magnitude of the Earth-frame velocity error and the magnitude of the ECEF position error of the iNavFIter algorithm over the first update interval. The attitude computation reaches the convergence criterion in 7 iterations, in contrast to the velocity/position computation in 5 iterations. The behaviors of attitude angle error and quaternion norm error are consistent with those observed in the QuatFIter [30], namely, having a non-uniform convergence rate with respect to iterations. Within four or five iterations, the quaternion norm error reduces to about $10^{-14}$, which helps us spare the quaternion normalization in computing the transformed specific force in (33). Otherwise, an extra approximation with Chebyshev polynomials has to be incorporated, as having been done for the gravity in (35). The gravity error in magnitude is presented in Fig. 5 during the velocity/position iterative computation, in which the gravity approximation reaches a quite good result within four iterations. As discussed regarding the flowchart of iNavFIter in Fig. 2, we could first compute the position, followed by the velocity computation

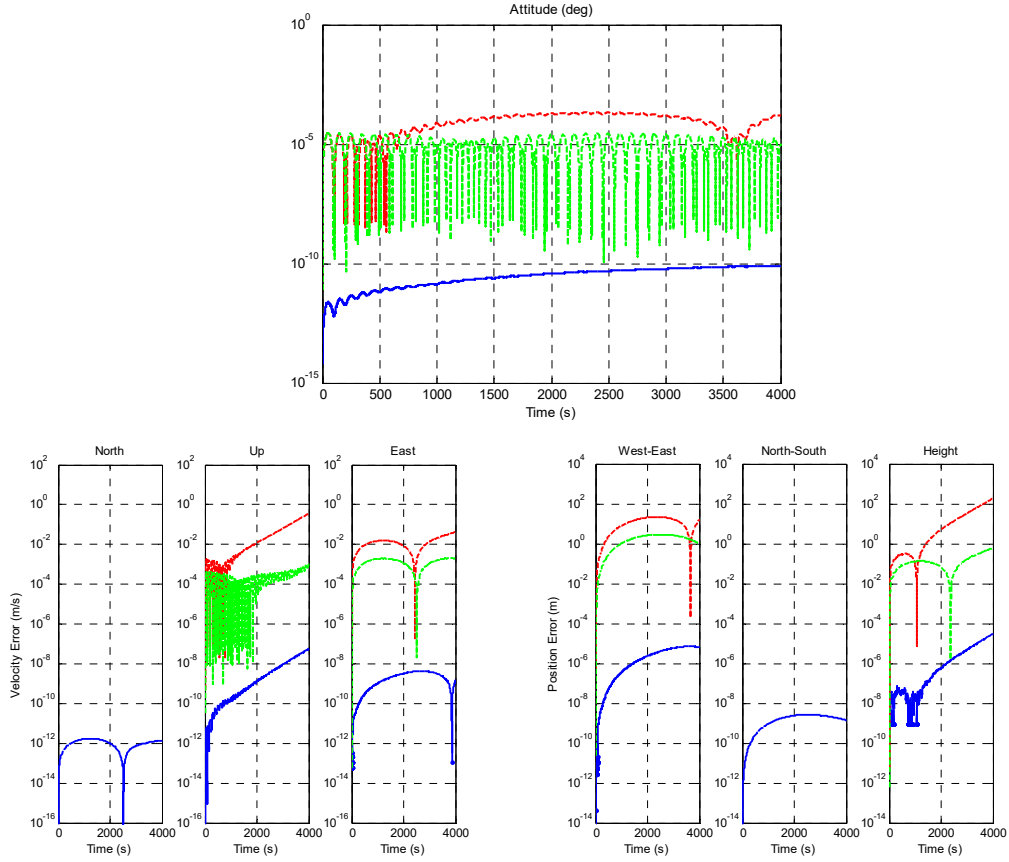

Figure 8. Navigation computation error of iNavFIter algorithm (blue solid line), typical 2-sample algorithm (red dashed line) and improved 2-sample algorithm considering reference frame rotation (green dashed line), for varying-speed level flight. Note that lines for north velocity error and north-south position error for 2-sample algorithms are absent because of zero values.

that uses the immediate position result, that is to say,

$$
\begin{aligned}
\mathbf{p}_{l+1}^{e} &= \mathbf{p}^{e}(0) + \frac{t_{N}}{2}\sum_{i=0}^{m_{v}}\mathbf{s}_{l,i}G_{i,[-1\ \tau]} \\
&= \sum_{i=0}^{m_{v}+1}\boldsymbol{\rho}_{l+1,i}F_{i}(\tau) \overset{\text{polynomial trucation}}{\approx} \sum_{i=0}^{m_{p}}\boldsymbol{\rho}_{l+1,i}F_{i}(\tau)
\end{aligned}
\tag{58}
$$

and

$$
\begin{aligned}
\mathbf{v}_{l+1}^{e} &= \mathbf{v}^{e}(0) + \frac{t_{N}}{2}\left(\mathbf{I}_{f} - 2\sum_{i=0}^{m_{v}}\boldsymbol{\omega}_{e}\times\mathbf{s}_{l,i}G_{i,[-1\ \tau]} + \sum_{i=0}^{m_{g}}\boldsymbol{\gamma}_{l+1,i}G_{i,[-1\ \tau]}\right) \\
&= \sum_{i=0}^{\max\{2m_{g}+n_{f},m_{v},m_{g}\}+1}\mathbf{s}_{l+1,i}F_{i}(\tau) \overset{\text{polynomial truncation}}{\approx} \sum_{i=0}^{m_{v}}\mathbf{s}_{l+1,i}F_{i}(\tau)
\end{aligned}
\tag{59}
$$

Note that in approximating the gravity vector in (59), the immediate position $\mathbf{p}_{l+1}^{e}$ (corresponding to $\boldsymbol{\gamma}_{l+1,i}$) is used instead of the position $\mathbf{p}_{l}^{e}$ in the last iteration. Magnitudes of the Earth-frame velocity error, position error and gravity error over the first update interval are plotted in Fig. 6 to show the consequence of the update order change and immediate result usage. As compared with Figs. 4-5, the velocity/position computation reaches the convergence criterion in 5 iterations as well, but the velocity, position and gravity all arrive at the converging states in fewer iterations (2 iterations), with the final velocity/position/gravity computation accuracy comparable with that in Figs. 4-5.

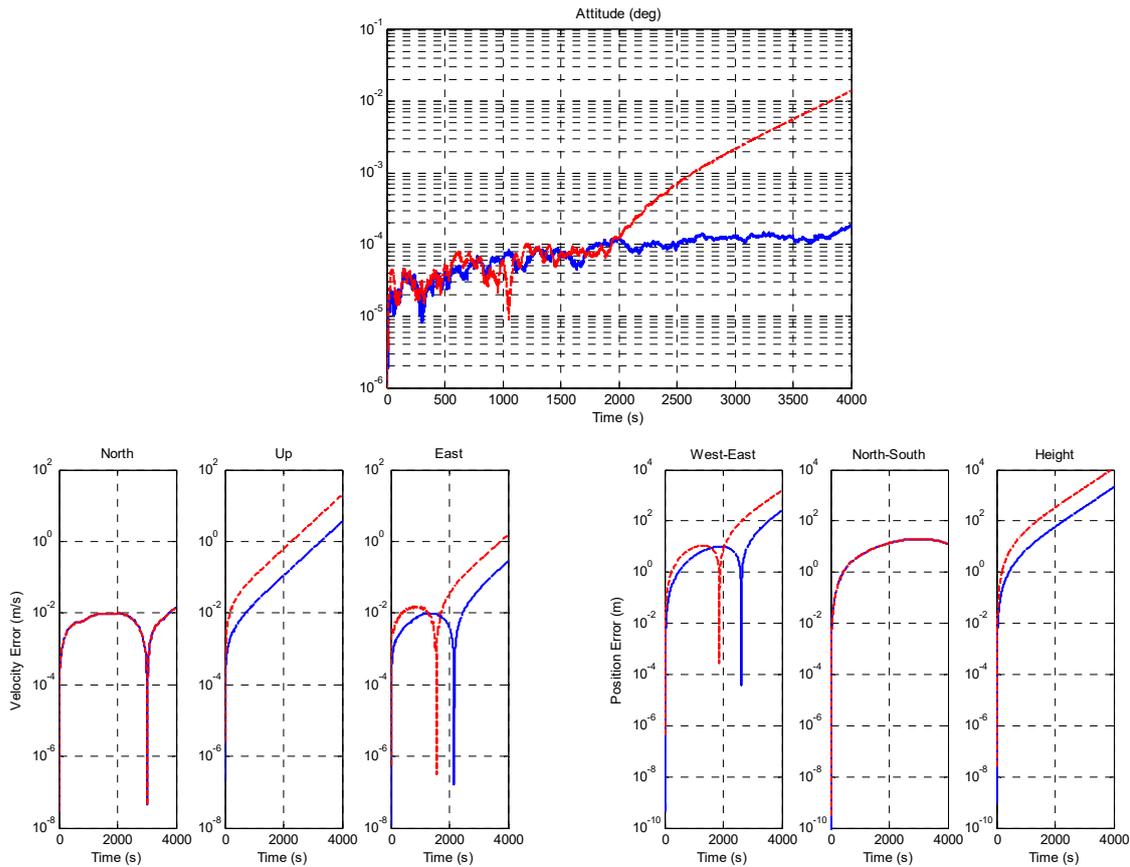

Figure 9. Navigation computation error of iNavFIter algorithm (blue solid line) and typical 2-sample algorithm (red dashed line) for simulated navigation-grade sensor during varying-speed coning flight.

Figure 7 compares the iNavFIter algorithm with the typical 2-sample algorithm [4, 5, 9] in navigation error for 4000 seconds. The comparison is performed in the local-level frame and any transformation error involved is unfairly owed to the iNavFIter algorithm. For instance, the coordinate transformation error from the ECEF position to the curvilinear position and then back to the ECEF position is about 3 nm [48]. We see that the iNavFIter algorithm performs tremendously better than the typical 2-sample algorithm by about 8~9 orders in attitude, velocity and position. As far as the west-east position error is concerned, for example, the iNavFIter algorithm's final error is 4 micrometers, in contrast to about 1200 meters of the typical 2-sample algorithm. It can be reasonably stated that the non-commutativity error, namely the well-known coning/sculling/scrolling errors, has been completely mitigated to "zero" or the machine precision.

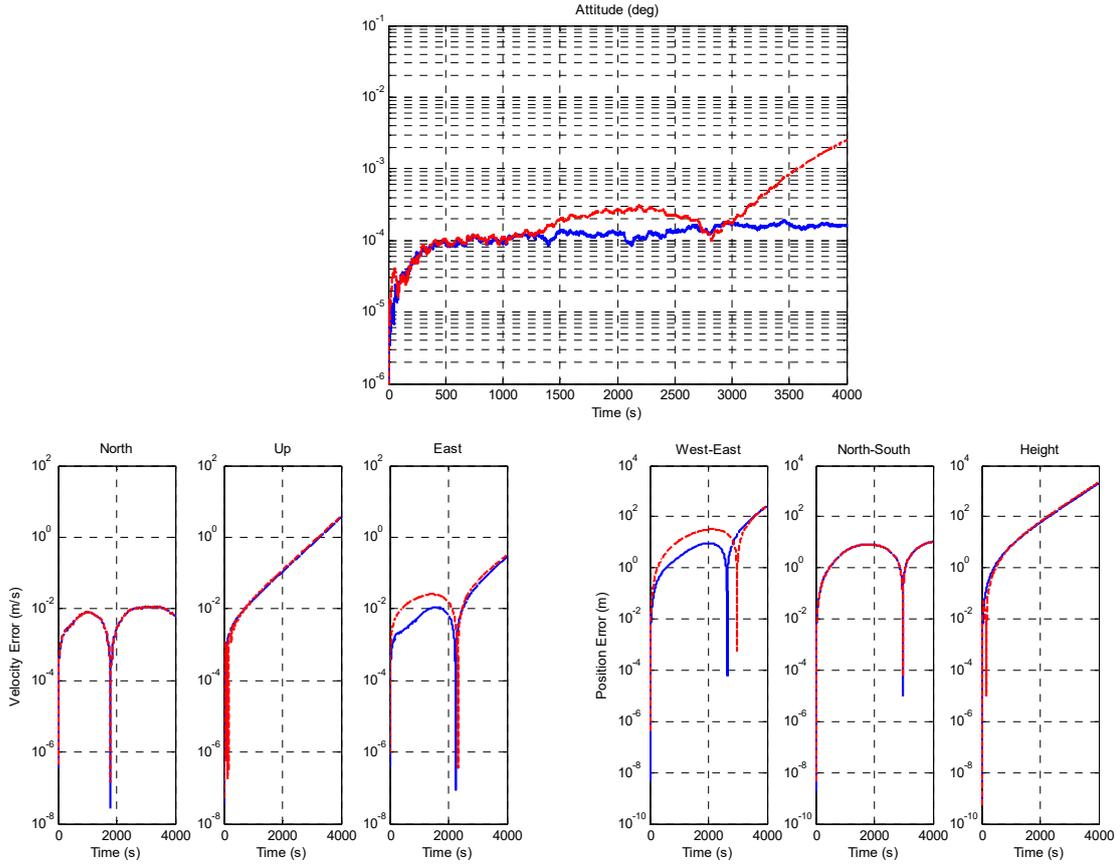

Figure 10. Navigation computation error of iNavFIter algorithm (blue solid line) and typical 2-sample algorithm (red dashed line) for simulated navigation-grade sensor during varying-speed level flight.

In order to make a peer comparison with the work [8], a level-flight case is next considered by letting $\mathbf{q}_n^b = \begin{bmatrix} 1 & 0 & 0 & 0 \end{bmatrix}^T$ so that the body frame coincides with the local-level frame during the whole flight. This case was designed therein to demonstrate the computation reference frame rotation-induced algorithm errors. Figure 8 plots the computation errors for the iNavFIter, the typical 2-sample algorithm [4, 5, 9], as well as the improved 2-sample algorithm [8] that tackles the rotation of the reference frame during the update interval. Contrasting Figs. 7-8, the typical 2-sample algorithm's error reduces nearly by two orders because it does not suffer from the body coning motion in the level-flight case. The further reduction of about one order achieved by the improved 2-sample algorithm is owed to the delicate treatment of the reference rotation frame in [8]. Additionally, the proposed iNavFIter demonstrates consistent excellency with almost the same accuracy in both cases of Figs. 7-8, which shows from another viewpoint that the attitude coning error and its effect on the subsequent velocity/position computation (in Fig. 7) has been substantially depressed.

In order to examine the iNavFIter's practical potential for the state-of-the-art inertial navigation systems, such as those based on accurate optical gyroscopes [1], sensor errors comparable to a high-end navigation-grade inertial navigation system are added to the true gyroscope/accelerometer outputs. Specifically, the bias and noise characteristics of the gyroscope triad are set to $10^{-4}$ deg/h and $10^{-4}$ $\mathrm{deg}/\sqrt{h}$ ; the bias and noise characteristics of the accelerometer triad are set to $10^{-5}$ m/s² and $10^{-6}$ $m/s^2/\sqrt{hz}$ . As shown in Fig. 9 for the coning-flight case, the computation error of the iNavFIter is still considerably smaller than that of the typical 2-sample algorithm, about 2 orders smaller in attitude error and about 6 times smaller in velocity and position errors. Figure 10 presents a result for the case of level flight, in which the iNavFIter is about 1 order smaller in the attitude error and about 2-3 times smaller in the east velocity and west-east position errors.

Table III summarizes the maximum west-east position errors of Figs. 7-10. It clearly shows (by contrasting Fig.7 with Fig. 8 and Fig. 9 with Fig. 10) that the attitude coning motion leads to a west-east position error of approximate 1200m in the typical 2-sample algorithm, which has been sufficiently depressed by the iNavFIter, in both scenarios of perfect sensor and navigation-grade sensor.

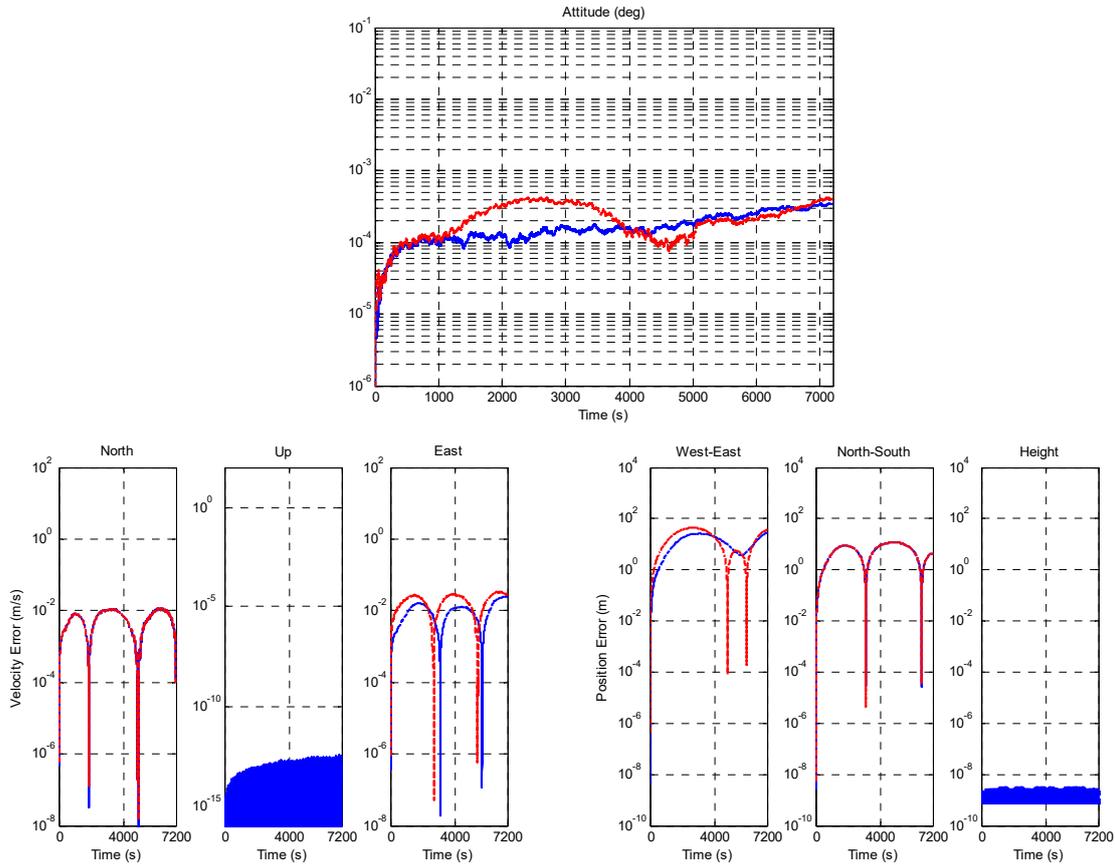

Figure 11. Navigation computation error of iNavFIter algorithm (blue solid line) and typical 2-sample algorithm (red dashed line) for simulated navigation-grade sensor during varying-speed coning flight, damped by zero vertical velocity and zero height. Note that lines for north velocity error and north-south position error for 2-sample algorithm are absent because of zero values.

Table III. Summary of Maximum West-East Position Errors in Simulated Flight Tests

| | Sensor Quality | Flight Type | iNavFIter | Typical 2-sample Algorithm |
|---|---|---|---|---|
| Fig. 7 | perfect sensors | coning flight | 4 μm | 1260 m |
| Fig. 8 | | level flight | 7 μm | 20 m |
| Fig. 9 | navigation-grade sensors | coning flight | 249 m | 1510 m |
| Fig. 10 | | level flight | 249 m | 266 m |

Comparing Fig. 7 with Fig. 9 and Fig. 8 with Fig. 10 indicates that the sensor imperfection results in a west-east position error of approximate 250m for both algorithms. The motion-incurred error component is all that the iNavFIter or any navigation algorithm can deal with.

The rising attitude errors of the 2-sample algorithm (after 2000s in Fig. 9 and after 2800s in Fig. 10) are owed to the coupling effect from the velocity/position to the attitude through $\boldsymbol{\omega}_{in}^{n}$ in the local-level frame mechanization (see Fig. 1), which actually results in the reference deviation of the local-level frame. A common practice is using a technique of vertical channel damping if the vertical velocity (such as in flat ground situations) or height information (by airborne barometers or underwater piezometers) is available. A bit straying from the focus of the current paper, the test is re-run with an additional damping step, i.e., brutally setting zero vertical velocity and zero height after each update interval in both algorithms. Figure 11 indicates that the damping technique, by incorporating a priori motion information, helps reduce the navigation errors of the 2-sample algorithm, as well as the velocity and position errors of the iNavFIter. Figure 12 demonstrates the navigation errors for the case of level flight as well. The west-east position errors of the iNavFIter and the 2-sample algorithm are, respectively, 28m and 27m in Fig. 11, and 44m and 50m in Fig. 12. The accuracy superiority of the iNavFIter in the east-west velocity/position errors reduces from over 6 orders (see Fig. 8) to less than one time. That the attitude errors of both damped algorithms for the level flight case (Fig. 12) are respectively comparable to those for the angular motion case (Fig. 11) indicates that for the considered simulation scenarios the attitude coning errors involved is negligible as compared with the attitude errors caused by the gyroscope imperfection, and the iNavFIter's accuracy superiority in velocity/position in Figs. 11-12 is largely owed to the delicate velocity/position computation. In other words, the first-order approximation in velocity/position computation contributes to the inferiority of the 2-sample algorithm. Inferred from the simulation observations so far, the iNavFIter's superiority is supposed to be more prominent for higher-quality inertial sensors. To conclude the simulation section, Figs. 13-14 present the damped algorithm results in 7200 seconds for further higher-quality gyroscopes (bias $10^{-5}$ deg/h and noise $10^{-5}$ $\deg/\sqrt{h}$) and accelerometers ($10^{-6}$ m/s$^2$ and $10^{-7}$ $m/s^2/\sqrt{hz}$). Table IV lists the maximum west-east position results for Figs. 11-14. As expected, the iNavFIter's superiority over the 2-sample algorithm increases up to about one order in the attitude error and east-west velocity/position errors.

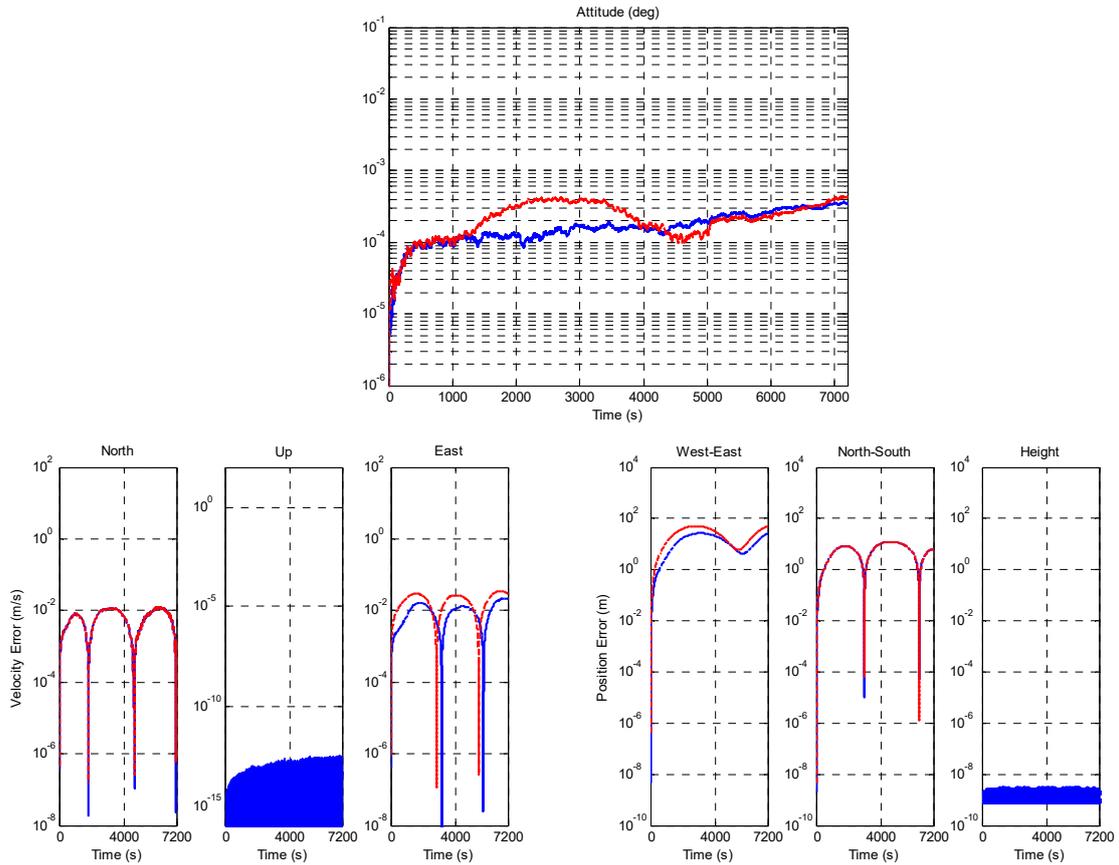

Figure 12. Navigation computation error of iNavFIter algorithm (blue solid line) and typical 2-sample algorithm (red dashed line) for simulated navigation-grade sensor during varying-speed level flight, damped by zero vertical velocity and zero height. Note that lines for north velocity error and north-south position error for 2-sample algorithm are absent because of zero values.

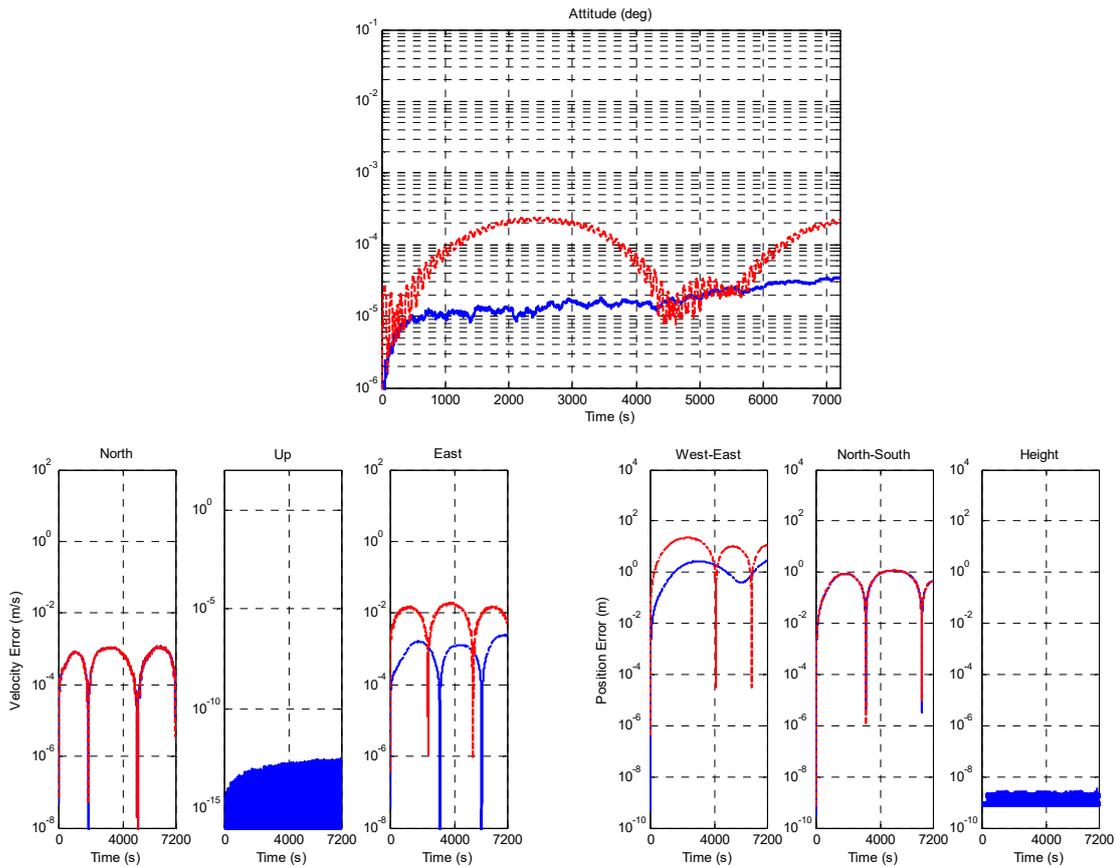

Figure 13. Navigation computation error of iNavFIter algorithm (blue solid line) and typical 2-sample algorithm (red dashed line) for simulated navigation-grade sensor (higher quality) during varying-speed coning flight, damped by zero vertical velocity and zero height. Note that lines for north velocity error and north-south position error for 2-sample algorithm are absent because of zero values.

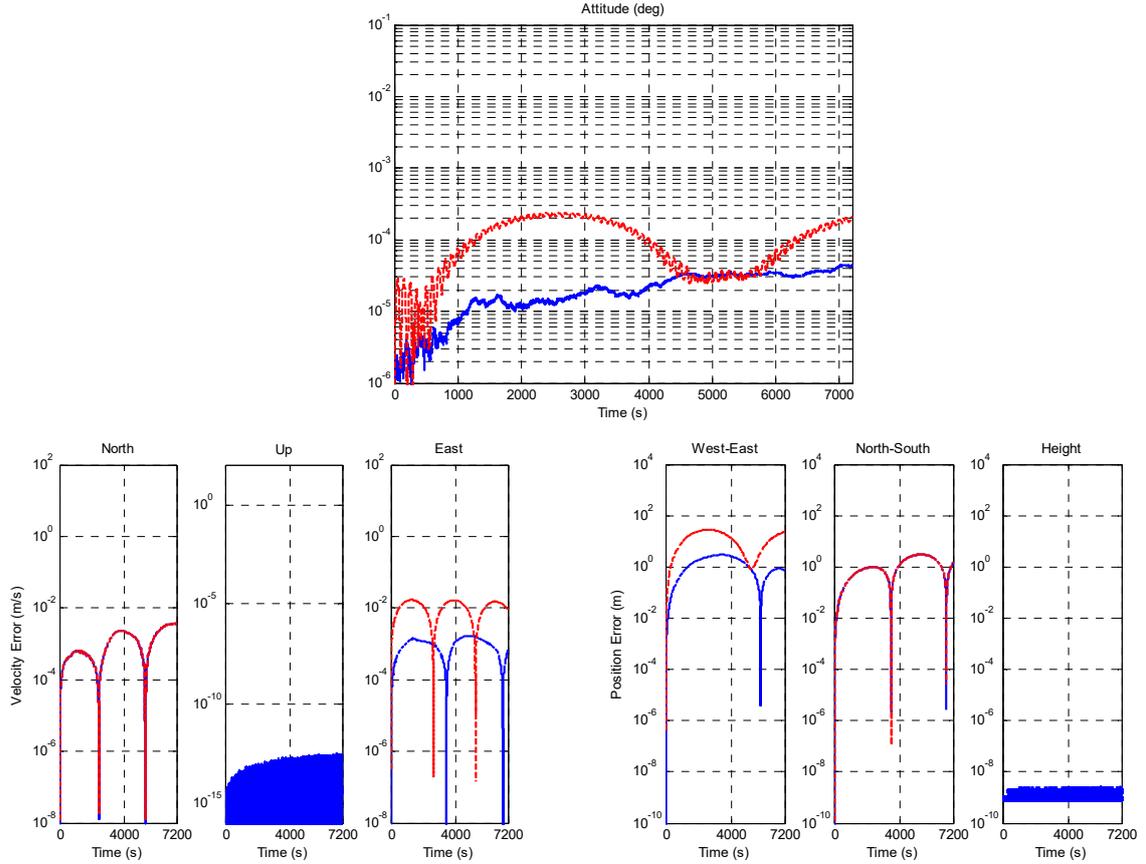

Figure 14. Navigation computation error of iNavFIter algorithm (blue solid line) and typical 2-sample algorithm (red dashed line) for simulated navigation-grade sensor (higher quality) during varying-speed level flight, damped by zero vertical velocity and zero height. Note that lines for north velocity error and north-south position error for 2-sample algorithm are absent because of zero values.

Table IV. Summary of Maximum West-East Position Errors by Damping Technique

|  | Sensor Quality | Flight Type | iNavFIter (*Damped*) | Typical 2-sample Algorithm (*Damped*) |
|---|---|---|---|---|
| Fig. 11 | navigation-grade sensors | coning flight | 28 m | 44 m |
| Fig. 12 | | level flight | 27 m | 50 m |
| Fig. 13 | higher-quality navigation-grade sensors | coning flight | 3 m | 22 m |
| Fig. 14 | | level flight | 3 m | 28 m |

All of the above numerical tests are performed on the Matlab platform. In terms of running time, the computational burden of the implemented iNavFIter is about ten times of that of the typical 2-sample algorithm, which is not a problem for modern computers or customized hardware implementation. The flight test datasets generated in this paper for both the coning-flight and level-flight cases are available online at https://www.researchgate.net/profile/Yuanxin_Wu/projects, under the project of 'Motion Representation and Computation - Inertial Navigation and Beyond'. Interested readers are free to test on them or their own data.

## VII. Conclusions and Discussions

Acquiring the attitude, velocity and position information is fundamental to motion body manipulation. This paper briefly reviews the development history and highlights two essential shortcomings of the strapdown inertial navigation algorithms, namely the theoretical simplification/approximation and the algorithmic design under special motion forms. The recent advance in precision attitude computation provides a promising method to surmount those shortcomings. Specifically, the polynomial fitting is applicable to any angular motion form and the functional iterative integration approach is capable of solving the full attitude kinematics with provable convergence. The Chebyshev polynomial can be used to speed up the implementation of the functional iterative integration yet at little expense of attitude accuracy.

In this paper, the functional iterative integration with Chebyshev polynomial approximation has been explored to tackle the whole process of inertial navigation computation, precisely integrating the kinematics of attitude, velocity and position fed on discrete inertial measurements by gyroscopes and accelerometers. The proposed iNavFIter algorithm chooses the Earth frame as the computation reference frame for the sake of computational efficiency, although other reference frames could also be used instead. For each update cycle, the iNavFIter algorithm is comprised of two consecutive iterative processes: attitude iterative computation and velocity/position iterative computation. Thanks to the loosely-coupling attributes of the Earth-frame mechanization, the attitude iterative computation runs independently and feeds the final iteration result as the input to the subsequent velocity/position computation. The velocity and position have to be iteratively computed together because of their mutual dependence on each other. Convergence and error analyses show that the iNavFIter has guaranteed convergence property under moderate situations. Numerical tests with analytically-generated trajectory datasets demonstrate the astonishing accuracy superiority of the iNavFIter algorithm over the state-of-the-art navigation algorithms. The iNavFIter actually brings the non-commutativity errors down to machine precision at affordable computation cost, namely the well-known coning/sculling/scrolling errors in attitude/velocity/position computation that have perplexed the navigation community for over half a century. This work is believed having paved a solid algorithmic road for the forthcoming ultra-precision inertial navigation system with meters-level or higher position accuracy, and the existing dynamic applications as well.

ACKNOWLEDGEMENTS

Special thanks to Dr. Qi Cai for Matlab code optimization.

APPENDIX. SOMIGLIANA GRAVITY FORMULA

The Somigliana gravity formula [52] is employed in this paper

$$g(L,h) = \gamma_e \frac{1 + k \sin^2 L}{\sqrt{1 - e^2 \sin^2 L}} \left[ 1 - \frac{2h}{R} \left( 1 + f + m - 2f \sin^2 L \right) + \frac{3h^2}{R^2} \right] \tag{60}$$

The parameter $k = r\gamma_p / R\gamma_e - 1$, where $\gamma_e$ and $\gamma_p$ are theoretical gravities at the equator and poles, $R$ and $r$ are semi-major and semi-minor axes of the ellipsoid, respectively, $e$ denotes the first ellipsoidal eccentricity, $f$ denotes the ellipsoidal flattening and $m = \Omega^2 R^2 r / GM$ where $GM$ is the Earth's gravitational constant.

REFERENCES

[1] J. F. Wagner and M. Perlmutter, "The ISS Symposium Turns 50: Trends and Developments of Inertial Technology during Five Decades," Karlsruhe, Germany, , 2015.

[2] P. G. Savage, "Blazing Gyros: The Evolution of Strapdown Inertial Navigation Technology for Aircraft," *Journal of Guidance, Control, and Dynamics,* vol. 36, pp. 637-656, 2013.

[3] D. A. Tazartes, "Inertial Navigation: From Gimbaled Platforms to Strapdown Sensors," *IEEE Trans. on Aerospace and Electronic Systems,* vol. 47, pp. 2292-2299, 2010.

[4] P. D. Groves, *Principles of GNSS, Inertial, and Multisensor Integrated Navigation Systems*, 2nd ed.: Artech House, Boston and London, 2013.

[5] D. H. Titterton and J. L. Weston, *Strapdown Inertial Navigation Technology*, 2nd ed.: the Institute of Electrical Engineers, London, United Kingdom, 2007.

[6] P. G. Savage, "Strapdown inertial navigation integration algorithm design, part 2: velocity and position algorithms," *Journal of Guidance, Control, and Dynamics,* vol. 21, pp. 208-221, 1998.

[7] P. Savage, "Down-Summing Rotation Vectors For Strapdown Attitude Updating (SAI WBN-14019)," Strapdown Associates (http://strapdownassociates.com/Rotation%20Vector%20Down_Summing.pdf)  2017.7.

[8] Y. Wu and X. Pan, "Velocity/Position Integration Formula (II): Application to Strapdown Inertial Navigation Computation," *IEEE Trans. on Aerospace and Electronic Systems,* vol. 49, pp. 1024-1034, 2013.

[9] P. G. Savage, *Strapdown Analytics*, 2nd ed.: Strapdown Analysis, 2007.

[10] P. G. Savage, "Computational Elements for Strapdown Systems," Low Cost Navigation Sensors and Integration Technology, NATO RTO-EN-SET-116, 2008.

[11] J. W. Jordan, "An accurate strapdown direction cosine algorithm,"  NASA TN-D-5384, 1969.

[12] J. E. Bortz, "A new mathematical formulation for strapdown inertial navigation," *IEEE Transactions on Aerospace and Electronic Systems,* vol. 7, pp. 61-66, 1971.

[13] M. Wang, W. Wu, J. Wang, and X. Pan, "High-order attitude compensation in coning and rotation coexisting environment," *IEEE Trans. on Aerospace and Electronic Systems,* vol. 51, pp. 1178-1190, 2015.

[14] J. G. Lee, Y. J. Yoon, J. G. Mark, and D. A. Tazartes, "Extension of strapdown attitude algorithm for high-frequency base motion," *Journal of Guidance, Control, and Dynamics,* vol. 13, pp. 738-743, 1990.

[15] P. G. Savage, "Strapdown inertial navigation integration algorithm design, part 1: attitude algorithms," *Journal of Guidance, Control, and Dynamics,* vol. 21, pp. 19-28, 1998.

[16] J. G. Mark and D. A. Tazartes, "On sculling algorithms," in *the 3rd St. Petersburg International Conference on Integrated Navigation Systems*, Central Scientific and Research Institute "Elektropribor", St. Petersburg, Russia, 1996, pp. 22-26.

[17] Y. Wu, X. Hu, D. Hu, T. Li, and J. Lian, "Strapdown inertial navigation system algorithms based on dual quaternions," *IEEE Transactions on Aerospace and Electronic Systems,* vol. 41, pp. 110-132, 2005.

[18] P. G. Savage, "A unified mathematical framework for strapdown algorithm design," *Journal of Guidance Control and Dynamics,* vol. 29, pp. 237-249, 2006.

[19] Y. Wu, "On "A unified mathematical framework for strapdown algorithm design"," *Journal of Guidance, Control, and*

*Dynamics,* vol. 29, pp. 1482-1484, 2006.

[20] Y. Wu and Z. Xiao, "On Position Translation Vector," in *AIAA Guidance, Navigation and Control Conference (fulltext available at https://arxiv.org/pdf/1112.5283)*, Minneapolis, Minnesota 2012.

[21] P. G. Savage, "Reply by the Author to Y. Wu et al.," *Journal of Guidance, Control, and Dynamics,* vol. 29, pp. 1485-1485, 2006.

[22] A. Soloviev, "Investigation into performance enhancement of integrated global positioning/inertial navigation systems by frequency domain implementation of inertial computational procedures," PhD. dissertation, College of Engineering and Technology, PhD thesis, Ohio University, 2002.

[23] Y. A. Litmanovich, V. M. Lesyuchevsky, and V. Z. Gusinsky, "Two new classes of strapdown navigation algorithms," *Journal of Guidance, Control, and Dynamics,* vol. 23, pp. 34-44, 28-30, Jun. 2000.

[24] J. G. Mark and D. A. Tazartes, "Tuning of coning algorithm to gyro data frequency response characteristics," *Journal of Guidance, Control, and Dynamics,* vol. 24, pp. 641-647, 2001.

[25] M. B. Ignagni, "Optimal strapdown attitude integration algorithms," *Journal of Guidance, Control, and Dynamics,* vol. 13, pp. 363-369, 1990.

[26] D. A. Tazartes and J. G. Mark, "Coning compensation in strapdown inertial navigation systems," US Patent US005828980A, 1997.

[27] Y. Wu, "RodFIter: Attitude Reconstruction from Inertial Measurement by Functional Iteration," *IEEE Trans. on Aerospace and Electronic Systems,* vol. 54, pp. 2131-2142, 2018.

[28] Y. Wu, Q. Cai, and T.-K. Truong, "Fast RodFIter for Attitude Reconstruction from Inertial Measurement," *IEEE Trans. on Aerospace and Electronic Systems,* vol. 55, pp. 419-428, 2019.

[29] G. Yan, J. Weng, X. Yang, and Y. Qin, "An Accurate Numerical Solution for Strapdown Attitude Algorithm based on Picard iteration," *Journal of Astronautics,* vol. 38, pp. 65-71, 2017.

[30] Y. Wu and G. Yan, "Attitude Reconstruction from Inertial Measurements: QuatFIter and Its Comparison with RodFIter," *IEEE Trans. on Aerospace and Electronic Systems (in press: https://ieeexplore.ieee.org/document/8686051),* 2019.

[31] M. Wang, W. Wu, X. He, G. Yang, and H. Yu, "Higher-order Rotation Vector Attitude Updating Algorithm," *Journal of Navigation,* vol. 72, pp. 721-740, 2019.

[32] Z. Xu, J. Xie, Z. Zhou, J. Zhao, and Z. Xu, "Accurate Direct Strapdown Direction Cosine Algorithm," *to appear in IEEE Trans. on Aerospace and Electronic Systems (early access: https://ieeexplore.ieee.org/document/8534467),* 2018.

[33] V. N. Branets and I. P. Shmyglevsky, *Introduction to the Theory of Strapdown Inertial Navigation System*: Moscow, Nauka (in Russian), 1992.

[34] C. Rucker, "Integrating Rotations Using Nonunit Quaternions," *IEEE Robotics and Automation Letters,* vol. 3, pp. 2779-2986, 2018.

[35] J. Park and W.-K. Chung, "Geometric integration on euclidean group with application to articulated multibody systems," *IEEE Trans. on Robotics,* vol. 21, pp. 850-863, 2005.

[36] M. S. Andrle and J. L. Crassidis, "Geometric Integration of Quaternions," *Journal of Guidance, Control, and Dynamics,* vol. 36, pp. 1762-1767, 2013.

[37] M. Boyle, "The Integration of Angular Velocity," *Advances in Applied Clifford Algebras,* vol. 27, pp. 2345–2374, 2017.

[38] P. Krysl and L. Endres, "Explicit Newmark/Verlet algorithm for time integration of the rotational dynamics of rigid bodies," *International Journal for Numerical Methods in Engineering,* vol. 62, pp. 2154–2177, 2005.

[39] E. Hairer, C. Lubich, and G. Wanner, *Geometric Numerical Integration: Structure Preserving Algorithms for Ordinary Differential Equations*. New York, NY, USA: Springer-Verlag, 2006.

[40] C. W. Clenshaw and H. J. Norton, "The Solution of Nonlinear Ordinary Differential Equations in Chebyshev Series," *Computer Journal,* vol. 6, pp. 88-92, 1963.

[41] J. Shaver, "Formulation and Evaluation of Parallel Algorithms for the Orbit Determination Problem," Ph.D., Dept. of Aeronautics and Astronautics, Massachusetts Inst. of Technology, Cambridge, MA, 1980.

[42] T. Fukushitma, "Picard Iteration Method, Chebyshev Polynomial Approximation, and Global Numerical Integration of Dynamical Motions," *The Astronomical Journal,* vol. 113, pp. 1909-1914, 1997.

[43] X. Bai and J. L. Junkins, "Modified Chebyhev-Picard iteration methods for orbit propagation," *The Journal of the Astronautical Sciences,* vol. 58, pp. 583-613, 2011.

[44] X. Bai, "Modified Chebyshev-Picard Iteration Methods for Solution of Initial Value and Boundary Value Problems," Ph.D. Dissertation, Texas A&M University, College Station, TX, 2010.

[45] Y. Wu, H. Zhang, M. Wu, X. Hu, and D. Hu, "Observability of SINS Alignment: A Global Perspective," *IEEE Trans. on Aerospace and Electronic Systems,* vol. 48, pp. 78-102, 2012.

[46] Y. Wu, C. He, and G. Liu, "On Inertial Navigation and Attitude Initialization in Polar Areas," *to appear in Satellite Navigation (online available at: https://arxiv.org/abs/1903.11863),* 2019.

[47] W. H. Press, *Numerical Recipes: the Art of Scientific Computing*, 3rd ed. Cambridge ; New York: Cambridge University Press, 2007.

[48] Y. Wu, P. Wang, and X. Hu, "Algorithm of Earth-centered Earth-fixed coordinates to geodetic coordinates," *IEEE Transactions on Aerospace and Electronic Systems,* vol. 39, pp. 1457-1461, 2003.

[49] W. E. Featherstone and S. J. Claessens, "Closed-form transformation between geodetic and ellipsoidal coordinates," *STUDIA*

*GEOPHYSICA ET GEODAETICA,* vol. 52, pp. 1-18, 2008.

[50] P. Civicioglu, "Transforming geocentric cartesian coordinates to geodetic coordinates by using differential search algorithm," *COMPUTERS & GEOSCIENCES,* vol. 46, pp. 229-247, 2012.

[51] G. Yan, W. Yan, and D. Xu, "Limitations of error estimation for classic coning compensation algorithm," *Journal of Chinese Inertial Technology,* vol. 16, pp. 380-385, 2007.

[52] *WGS 84 TECHNICAL REPORT - NIMA TR8350.2 Department of Defense World Geodetic System 1984, Its Definition and Relationships With Local Geodetic Systems (Third Edition ed.).* Available: http://earth-info.nga.mil/GandG/wgs84/gravitymod/